%% file: main.tex
\documentclass[11pt, a4paper, logo, copyright, nonumbering]{lti}
\usepackage[authoryear, sort&compress, round]{natbib}
\usepackage{dblfloatfix}
\usepackage{ulem}
\usepackage{caption}
\usepackage{dramatist}
\usepackage{xspace}
\usepackage{pifont} 
\usepackage{multirow}
\usepackage{tcolorbox}
\usepackage{xltabular}
\usepackage{longtable}
\usepackage{hyperref}
\usepackage{microtype}
\usepackage{amsfonts}
\usepackage{amsmath}
\usepackage{amssymb}
\usepackage{lineno}
\usepackage{multirow}
\usepackage{adjustbox}

\usepackage[bottom]{footmisc}

\usepackage{CJKutf8}
\usepackage{subfigure}
\usepackage{setspace}

\usepackage{dsfont}
\usepackage{array}
\usepackage{tabularx}
\usepackage{subfigure}
\usepackage{xcolor}
\usepackage{wrapfig}
\usepackage{booktabs}

\definecolor{ltired}{HTML}{780000}

\usepackage{lipsum}
\usepackage{multicol}
\usepackage{makecell}
\usepackage{algorithm}
\usepackage{algorithmicx}
\usepackage{algpseudocode}

\newcommand{\diyi}[1]{\ifthenelse{\boolean{showcomments}}{\textcolor{blue}{[#1 —diyi]}}{}}
\newcommand{\liwei}[1]{\ifthenelse{\boolean{showcomments}}{\textcolor{orange}{[#1 —Liwei]}}{}}
\makeatletter
\def\@BTrule[#1]{%
  \ifx\longtable\undefined
    \let\@BTswitch\@BTnormal
  \else\ifx\hline\LT@hline
    \nobreak
    \let\@BTswitch\@BLTrule
  \else
     \let\@BTswitch\@BTnormal
  \fi\fi
  \global\@thisrulewidth=#1\relax
  \ifnum\@thisruleclass=\tw@\vskip\@aboverulesep\else
  \ifnum\@lastruleclass=\z@\vskip\@aboverulesep\else
  \ifnum\@lastruleclass=\@ne\vskip\doublerulesep\fi\fi\fi
  \@BTswitch}
\makeatother

\addto\extrasenglish{
}

 {\begin{list}{}%
         {\setlength{\leftmargin}{#1}}%
         \item[]%
 }
 {\end{list}}
 
\newcommand{\cmu}{\raisebox{-2.5mm}{\includegraphics[width=3.5mm]{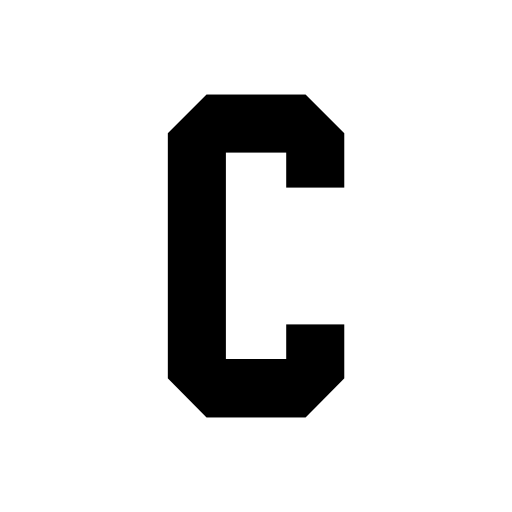}}}
\newcommand{\uw}{\raisebox{-2.6mm}{\includegraphics[width=2.8mm]{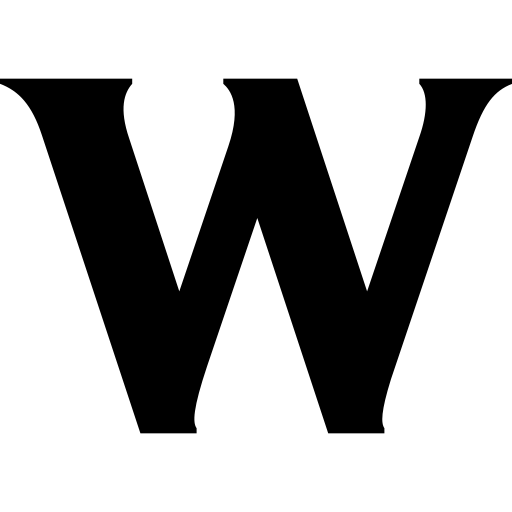}}}
\newcommand{\hs}{\raisebox{-2.6mm}{\includegraphics[width=2.0mm]{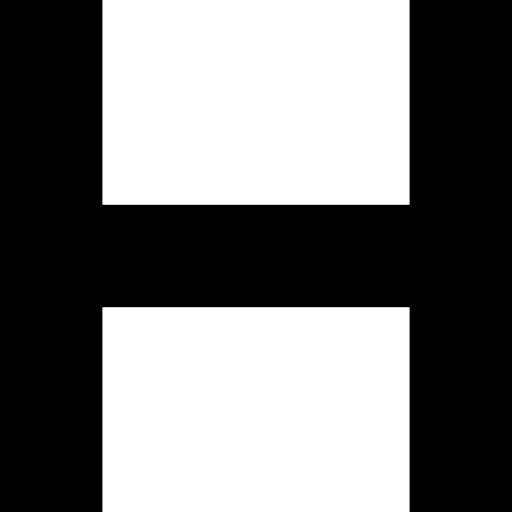}}}
\newcommand{\ucsd}{\raisebox{-2.6mm}{\includegraphics[width=2.0mm]{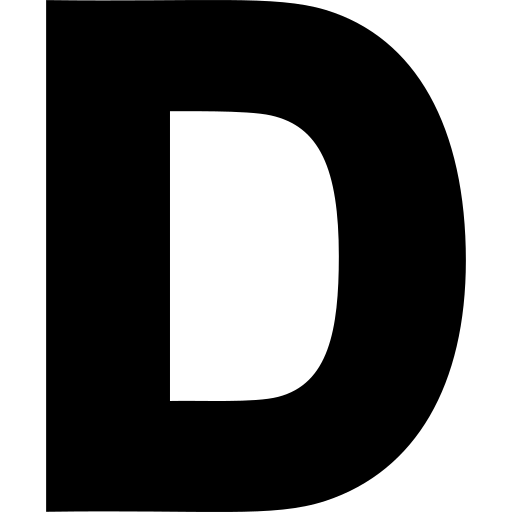}}}
\newcommand{\stfd}{\raisebox{-2.6mm}{\includegraphics[width=2.0mm]{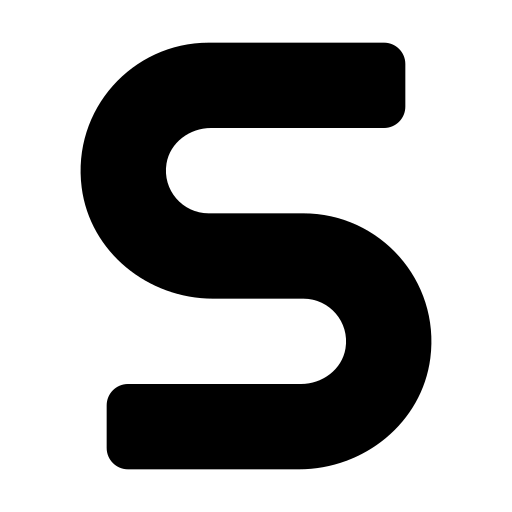}}}
\newcommand{\princeton}{\raisebox{-2.6mm}{\includegraphics[width=2.0mm]{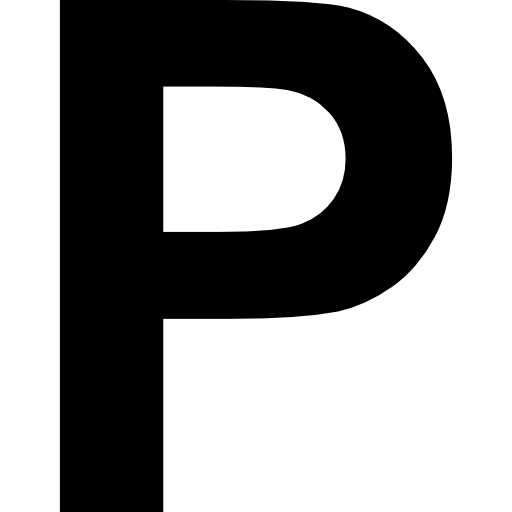}}}

\usepackage{color, colortbl}
\usepackage{color-edits}
\addauthor{df}{cyan}
\addauthor{gn}{magenta}
\addauthor{dy}{blue}
\addauthor{zw}{orange}
\addauthor{ag}{brown}
\addauthor{rs}{cyan}
\addauthor{pw}{violet}

\definecolor{amber}{rgb}{1.0, 0.75, 0.0}
\definecolor{applegreen}{rgb}{0.55, 0.71, 0.0}
\definecolor{treegreen}{rgb}{0.13, 0.54, 0.23}
\definecolor{LightCyan}{rgb}{0.88,1,1}

\title{\centering Efficient Test-Time Adaptation through \\Human-Agent Interaction}

\author{\centering
\normalsize{\textbf{
Zora Zhiruo Wang$^{\cmu}$ \quad Apurva Gandhi$^{\cmu}$ \quad Rulin Shao$^{\uw}$ \quad Aspen Chen$^{\cmu}$ \\ Jonas Mueller$^{~\hs}$ \quad {Zhiqi Liang}$^{\ucsd}$ \quad {Jett Chen}$^{\cmu}$ \quad {Michael Ryan}$^{\stfd}$ \quad {Qianou Ma}$^{\cmu}$ \\}}
\small{\textbf{
{Luxi He}$^{\princeton}$$~~${Zhoujun Cheng}$^{\ucsd}$$~~${Andre He}$^{\cmu}$$~~${Seungone Kim}$^{\cmu}$$~~${Jiayi Geng}$^{\cmu}$$~~${Mingqian Zheng}$^{\cmu}$ \\ {Weiwei Sun}$^{\cmu}$$~~${Zheyuan Zhang}$^{\cmu}$$~~${Xinran Zhao}$^{\cmu}$$~~${Yike Wang}$^{\uw}$$~~${Abe Hou}$^{\stfd}$$~~${Liwei Jiang}$^{\uw}$ \\}}
\normalsize{\textbf{
{Pang Wei Koh}$^{\uw}$ \quad {Diyi Yang}$^{\stfd}$ \quad {Graham Neubig}$^{\cmu}$ \quad {Daniel Fried}$^{\cmu}$
}} 
\\
\vspace{0.3em}
$^{\cmu}$Carnegie Mellon University \quad $^{\uw}$University of Washington \quad $^{\hs}$Handshake AI \\
$^{\stfd}$Stanford University \quad $^{\princeton}$Princeton University \quad $^{\ucsd}$University of California San Diego
\\
\vspace{0.3em}
\small{\texttt{zhiruow@cs.cmu.edu}}
}

\input{commands.tex}

\input{sections/0_abstract}

\begin{document}

\maketitle

\input{sections/1_introduction}
\input{sections/2_method}
\input{sections/3_expertise}
\input{sections/4_rubrics}

\input{sections/5_community}
\input{sections/discussions}

\newpage
\bibliography{main}

\newpage
\appendix
\input{appendix/2_problem_method}
\input{appendix/method-ablation}
\input{appendix/3_agent-adaptation}

\input{appendix/4_rubrics}
\input{appendix/lm-prompts}

\end{document}

%% file: commands.tex
\newcommand{\model}{\textsc{Model}}

\renewcommand{\phi}{\varphi}

\renewcommand{\epsilon}{\varepsilon}
\renewcommand{\imath}{\mathrm{i}}

\newlength{\restsubwidth}
\newlength{\restsubheight}
\newlength{\restsubmoreheight}
\newcommand{\rest}[2]{%
        \settowidth{\restsubwidth}{\ensuremath{#2}}
        \settoheight{\restsubheight}{\ensuremath{{}_{#2}}}
        \ensuremath{{#1\hskip 0.5pt}_{\vrule\kern2pt\parbox[b][%
        4pt][b]{\the\restsubwidth}{%
                        \ensuremath{{}_{#2}}}}}
        }

%% file: sections/0_abstract.tex
\begin{abstract}
AI agents are trained on population-scale data to encode broad capabilities spanning those of many practitioners. Yet the artifacts they produce rarely meet the personal bar professionals need to stake their reputation on. On realistic, open-ended tasks where success criteria are heterogeneous and insufficiently documented, individual expertise lives precisely in the elevation and departure from the average.
In practice, iterative human-agent interaction surfaces criteria that users cannot fully specify up front, yet apply repeatedly across tasks.
We argue this cross-session interaction data is a rich, underused signal for closing the gap to individual expertise.
In this work, we propose \underline{t}est-time \underline{a}daptation through \underline{h}uman-agent \underline{i}nteraction (\textsc{TAHI}), which integrates these signals into agent context and weights, and crystallizes each user's training and evaluation criteria via an evolving rubric module.
We adapt agents to 30 individuals in two high-utility domains, writing and visual creation, on a total of 600 tasks. Our agents improve on solo task success by 4.5--20.9\% within only tens of tasks. Meanwhile, our evolving rubric module serves as a scalable annotation tool, creating evaluation rubrics that catch 16.0--22.3\% more failures than those from LMs or humans alone.
While agents are adapted towards individuals, we show these personalized agents also produce improvements in success of up to 8.8\% that generalize across users.\footnote{}
\end{abstract}

%% file: sections/1_introduction.tex
\section{Introduction}

Large language model (LLM) based agents have rapidly matured into capable generalists, producing reasonable outputs across writing, software engineering \citep{wang2025how}, and a growing range of occupational tasks \citep{patwardhan2026gdpval}.
Yet this general competence still falls short of expert standards in notable ways: flattening responses into homogeneity \citep{jiang2026artificial} or drifting away from users' original opinions \citep{abdulhai2026llms}, yielding outcomes that rarely meet the bar for professionals to use without further iterative editing.

Human expertise, particularly the tacit practice and judgment accumulated over years of domain experience, is only partially captured by the population-scale textual corpora used to train foundational LMs \citep{Gobet1998ExpertMA}. A model necessarily learns broad capabilities spanning many practitioners, but an individual professional needs specific knowledge, judgment, and strategies that elevate and depart from the population average.
This tension only sharpens on realistic, open-ended tasks, which have diverse success criteria that are seldom fully documented \citep{weidinger2025toward}, leaving a constant gap between the capabilities an agent can learn from documented human knowledge and those that remain unverbalized. Compounding this, expertise is inherently individual and can conflict across practitioners, yielding a set of desiderata that is hard for a single trained model or default agent framework to satisfy simultaneously.
Closing this last mile, then, requires agents to adapt at deployment time, on the fly, to the individuals they are working with \citep{kojima2021continual,hawkins2020continual}.
Moreover, it is important to do so efficiently, converging on a user's preferred way of working within a handful of sessions, before their patience for dozens of trial rounds runs out \citep{shaikh2025aligning}.

\begin{figure*}[t!]
\centering
    \includegraphics[width=\textwidth]{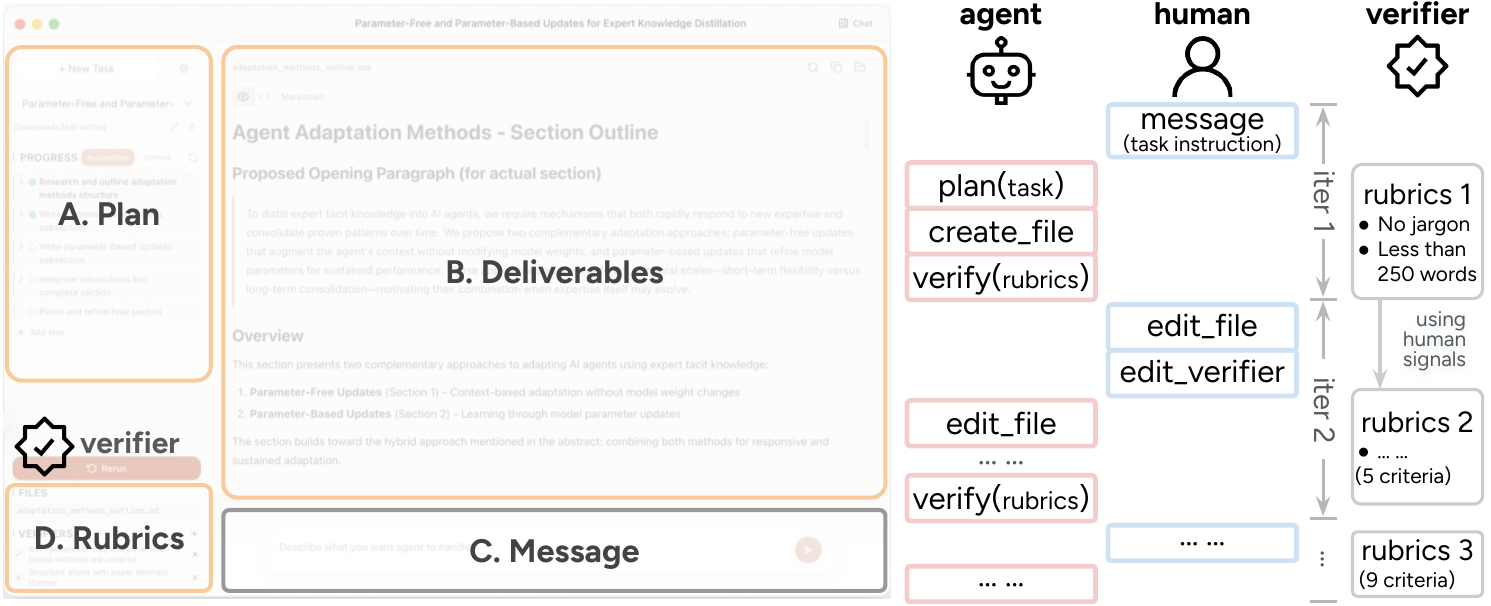}
    
    \vspace{4mm}
    \includegraphics[width=\textwidth]{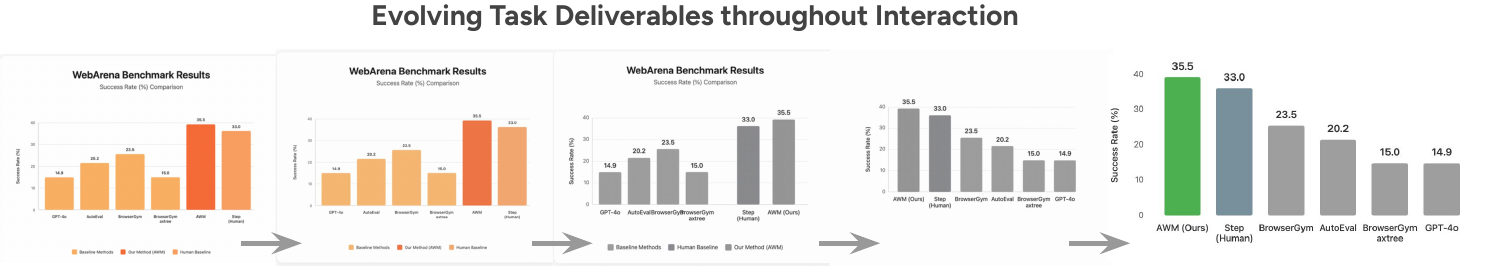}
\caption{Illustration of a human-agent interaction process using our interface with an evolving verifier module.
The agent executes tasks, while the human provides instructions, suggestions, and corrections via four modules: plan adjustment (A. Plan), output file editing (B. Deliverables), textual feedback (C. Message), and criteria specification (D. Rubrics).
Human and agent iterate on this instruction-execution loop to improve the deliverables (iter 1 $\rightarrow$ iter 2 $\rightarrow$ ...) until the human is satisfied with it (bottom). Meanwhile, the verifier module evolves by integrating human signals from the previous iteration, updating the rubrics to capture human preferences.
}
\label{fig:agent-interface}
\end{figure*}

Efficient test-time adaptation, however, demands the right learning signal.
When agents improve solely by reflecting on their past attempts, progress plateaus in the absence of new signals from the outside \citep{huang2024large,xu2024pride}.
It often takes many rounds of human-agent interaction to iterate toward a target shape the user could not fully specify up front. Many of these signals recur across task sessions \citep{wang2025agent}, yet current agent systems largely leave them unused when interacting with humans. It is precisely this multi-session interaction data, accumulated through the process of working with the agent, that offers a rich, under-explored learning signal for agents.
Unlike a one-off correction, we envision a human mentor working alongside a trainee agent across multiple task sessions, offering continual, hands-on suggestions across multiple channels. The challenge, then, is to collect and integrate these interaction streams into the agent.

To this end, we propose \underline{t}est-time \underline{a}daptation through \underline{h}uman-agent \underline{i}nteraction (\textsc{TAHI}).
In this framework, human interaction signals are processed in a streaming manner, where agents continually adapt themselves based on previous interaction sessions to solve subsequent sessions, progressively reducing the interactions needed to meet the human experts' expectations (\autoref{fig:agent-adaptation}).
Our agent is supported by a trainable backbone LM (Qwen3.6-35B) conditioned on editable contexts, and an interface that affords various channels of human interaction (\autoref{fig:agent-interface}). Notably, besides using human textual feedback \citep{song2026expanding} as in existing frameworks (our \textit{Message} module), our interface enables versatile additional signals, including planning adjustments (\textit{Plan}), verification specifications (\textit{Rubrics}), and direct edits to produce artifacts (\textit{Deliverables}); allowing us to study the value of versatile human signals.

Alongside the collection of human-agent interaction data, \textsc{TAHI} supports test-time agent adaptation and rubric creation for the open-ended tasks we evaluate on. 
For one, our agent allows (i) context adaptation, by inducing factual memory and procedural skills, that enables easy inspection and revision of agent learning by the user; and (ii) weight adaptation, by encouraging agents to produce human-guided final output over its initial one using direct preference optimization (DPO, \citep{rafailov2023direct}) on LoRA adapters \citep{hu2022lora}, to capture more persistent behavioral changes and spares the context window from overgrowing.
For another, we embed an evolving verifier module that uses an LLM to crystallize human activity into concrete evaluation criteria, while allowing users to correct anything that appears inaccurate. It serves as a scalable tool to annotate criteria that agents can be trained and evaluated against, and spares users the effort of hand-writing success criteria for open-ended tasks (\S\ref{sec:2:method}).

We individually adapt agents to 30 human experts across 600 tasks in two domains with wide AI use, namely writing and visual creation. For each domain, we collect 5 humans' interaction data with our agent in three scenarios: online context adaptation, online weight adaptation, and offline (no adaptation); with each human doing 20 tasks in total. From this data, we find:

\noindent $\bullet$ \textbf{Agents effectively and efficiently adapt to individual users.} 
Under both context and weight adaptation, our agents achieve similarly substantial improvements of 4.5--12.9\% and 4.5--20.9\% in solo task success rate, and do so efficiently within only 20 task sessions. On additional held-out tasks, our context- and weight-adaptive agents improve by 3.2--6.6\% and 4.8--5.2\% on rubrics tied to the same humans, signaling strong cross-task generalization capabilities (\S\ref{sec:3:expertise}).

\noindent $\bullet$ \textbf{Our evolved rubrics are more comprehensive than those created by LLMs or humans alone.} 
LLM-produced rubrics often miss critical evaluation aspects, and humans require substantial motivation and training to avoid the same pitfalls. Our evolved rubrics, in comparison, can capture 16.0--22.3\% more failures on imperfect agent solutions, compared to rubrics produced by LLMs and even humans alone (\S\ref{sec:4.1:rubric-comparison}).

\noindent $\bullet$ \textbf{Our individually adapted agents improve across different users.} 
Our data reflects that human expertise is a composition of shared community knowledge and personalized strategies or preferences. Our agents learn both, achieving 0.3--8.8\% and 6.2--19.6\% gains on shared and personal rubric items across users (\S\ref{sec:4.2:rubric-composition}-\S\ref{sec:4.4:consolidation}).

Looking forward, we envision a future where AI agents are not generic, but adaptive tools shaped by, and accountable to, the individuals they serve.

%% file: sections/2_method.tex
\section{Efficient Test-Time Adaptation through Human-AI Interaction}
\label{sec:2:method}

\subsection{Fundamentals: Human-Agent Interaction}

We formalize the problem as a human-agent interaction \citep{cukurova2026you} where, given a task specified by the human, the agent takes actions to execute it, while the human provides feedback and corrections along the way.

\paragraph{Modeling Agent and Human}
We define two parties involved in a task session. First, an LM-based agent $A$ operating with policy $\pi^A_{\theta}(\cdot | c)$ parameterized by an LM $\theta$ and conditioned on context $c$. Following two threads of work adapting agents in context, we design two components: (i) memory storing factual and preferential knowledge \citep{xu2026mem}, and (ii) skills storing procedural knowledge \citep{wang2024voyager,skills2025}. 
As our goal is for agents to adapt towards the human, we also explicitly define human $H$ operating with policy $\pi^H$.

Given a task with instruction $q$, both parties can take actions from their respective action space to form an action trajectory $\tau = [(s_0, o_0, a_0), (s_1, o_1, a_1), \cdots]$ that solves the task. $s_i$ represents the shared environment state, $o_i$ is the observation of the action-taking party (agent $A$ or human $H$), and $a_i$ is the action being taken.

For agents, we adopt the common action space for both GUI and coding activities. We adopt coding agents as generalist agents, taking the position that programmatic actions subsume common computer-use actions (e.g., \texttt{click}, \texttt{scroll}, \texttt{type}) and suffice for solving varied tasks.
$$\mathcal{A}^A = \{\texttt{message}, \texttt{file\_create}, \texttt{file\_edit}, \texttt{file\_read}, \texttt{execute}, \texttt{plan} \}$$
This action space endows the agent with the following capabilities. 
(1) communication (\texttt{message}): surfacing information or requesting clarification via messages;
(2) file manipulation (\texttt{file\_create}, \texttt{file\_edit}, \texttt{file\_read}): reading, creating, or editing files in the environment;
(3) program execution (\texttt{execute}): executing program scripts; and
(4) planning (\texttt{plan}): creating and refining structured step decomposition, as well as verification criteria for each step.

The human acts within a complementary action space: 
$$\mathcal{A}^H = \{\texttt{message}, \texttt{file\_edit}, \texttt{plan\_edit}, \texttt{context\_edit}, \texttt{verify}, \texttt{trigger} \}$$
Humans share a few actions similar to agents:
(1) communication (\texttt{message}): providing feedback or additional information to the agent via text messages;
(2) file manipulation (\texttt{file\_edit}): editing agent-produced artifacts;
(3) planning (\texttt{plan\_edit}): modifying the agent-proposed step decomposition and the verification criteria; 
Meanwhile, humans are equipped with more actions for supervising and adjusting the agent, via:
(4) verification (\texttt{verify}): providing pass/fail signals to verification criteria; 
(5) edit agent context (\texttt{context\_edit}): edit the textual content in memory and skills files; and
(6) agent solution refinement (\texttt{trigger}): signals the end of human activities and triggers the agent to act further.

The entire task session $\tau$ can be viewed as interleaved $J+1$ turns conducted by agent and human, where each turn is itself a trajectory of actions, $\tau^*_*$'s, composed of one or more actions. Formally, $\tau = [\tau^H_0, \tau^A_0, \tau^H_1, \tau^A_1, \cdots, \tau^H_J, \tau^A_J]$, where $\tau^H_0 = [\texttt{message}(q)]$ represents the user task instruction step, with $q$ being the text of instruction. We determine the end of a task session if the user does not follow up or take further actions. If $\tau^H_J$ triggers another round of agent actions (i.e., last action in $\tau^H_J$ is \texttt{trigger}), $\tau^A_J$ will be non-empty; otherwise $\tau^A_J = \emptyset$, indicating the user is satisfied with the solution after their own actions.

\paragraph{Agent Interface Modification to Encourage Versatile Human Signals}
To make it easy for human users to express versatile signals while preserving a smooth user experience, we build on top of the widely adopted open-source Agent Cowork\footnote{\url{https://github.com/DevAgentForge/Open-Claude-Cowork}} and add three components:

First, \textit{initial task planning} (\autoref{fig:agent-interface}, Plan). We ask the agent to propose an initial task plan, displayed in the interface Plan module, so that human users can give targeted feedback on it. Concretely, this planning action decomposes the instruction into multiple steps, each represented as a textual description $w^i$, and comes with an output artifact $f^i$. For each step, we design an LM-supported \textsc{verifier} module that initiates a list of NL verification criteria $V^i = \{v^i_k\}$ to judge how much $f^i$ satisfies $w^i$. We allow users to edit the steps, files, and verifiers by adjusting the \textit{Plan}, \textit{Files}, and \textit{Rubrics} modules in the interface.
Formally, this action can be denoted as $a^A_{0,0} = \texttt{plan}(o^A_{0,0}) \rightarrow \{(w^i, f^i, V^i)\}$, where $a^A_{0,0}$ denotes the first action taken in the first agent turn of trajectory $\tau^A_0 = [(s^A_{0,0}, o^A_{0,0}, a^A_{0,0}), \cdots]$. Since users mostly interact with the files and verifiers produced in the last step, for simplicity, we use $f$ and $V$ to denote the final artifacts and rubrics.
 
Second, \textit{direct file editing} (\autoref{fig:agent-interface}, Deliverables). Existing interfaces mostly require all human feedback to be verbalized through messages (C. Message), which is limiting when a desired change is hard for humans to verbalize accurately. We instead switch the main panel to be a display of agent-produced deliverables (e.g., written documents, data visualizations), allowing users to directly edit them. 
Since humans favor UI-based interactions despite agents often operating in programmatic ways \citep{wang2025how}, for visual deliverables (e.g., figures in HTML), our interface allows users to drag elements directly via the UI, which are translated into code diffs for the agent's observation \citep{cursor2026direct}.

Third, \textit{automatic step verification} (\autoref{fig:agent-interface}, Rubrics). 
The agent fulfills step goal $w^i$ by generating a trajectory $\tau^i$ to produce artifact $f^i$. The \textsc{verifier} module grades $f^i$ against the criteria $V^i$, producing a list of binary pass/fail scores as $\texttt{verify}(f^i, v^i_k) \rightarrow r^i_k \in \{0, 1\}$. 
The grading outputs will be rendered on the interface as red crosses or green checks for the user's information. Users can change the grading if they disagree with the automatic verification result by executing the $\texttt{verify}$ action. It serves as an assistive tool for users to check intermediate task success, and lays the foundation for constructing the task evaluation rubrics.

\paragraph{The Iterative Nature of Task Sessions}
Particularly for artifact-driven tasks, the human-agent interleaving action-taking often happens for multiple rounds, to produce the final artifacts that satisfy the users. In response to this iterative nature of task sessions, we can view each pair of human-agent trajectories $(\tau^H_j, \tau^A_j)$ as an interaction unit $u_j$, producing the output artifact $f_j$ on top of the previous $j$ iterations producing $f_0, \cdots, f_{j-1}$.
Concretely, after each interaction unit $u_{j}$, the user can start another iteration by taking a sequence of actions forming $\tau^H_{j+1}$. 
Once the user finishes acting and triggers, the agent will take in all past interactions $u_{0:j}$ and the latest human activity $\tau^H_{j+1}$, to take actions $\tau^A_{j+1}$ and refine its own artifact output to $f_{j+1}$. 

While the automatic verification module produces a quality measure $r_0$ of the current artifact, the initial rubrics $V_0$ are created in one pass by an LM and may not capture all criteria for faithful evaluation. Therefore, we enable evolution of the \textsc{verifier} module to update the last-iteration criteria $V_{j}$ using human activities $\tau^H_j$, by verbalizing users' underlying intent and preferences into gradable rubrics, yielding updated rubrics $V_{j+1}$.
In an ideal case, the final evolved verifiers $V_J$ should capture all user requirements and serve as comprehensive evaluation rubrics, the quality of the resulting artifact $f_j$ should be roughly increasing throughout task iterations (i.e., as $j$ increases) by integrating human feedback, therefore $\sum_{v \in V_J} \texttt{verify}(f_{j'}, v) \ge \sum_{v \in V_J} \texttt{verify}(f_{j}, v), \forall j' > j$.
Nonetheless, in practice, we observe this to hold when $j$ and $j'$ are farther apart, e.g., when $j=0, j'=J$, which serves as a stronger agent learning signal.

\subsection{Test-Time Agent Adaptation Paradigm}
\label{sec:2.2:adapt-paradigm}

Test-time agent adaptation operates in a streaming manner, where tasks arrive sequentially and each completed session provides new data for adapting the agent. 
Unlike one-time adaptation from a fixed dataset, the agent continually updates as it accumulates experience with the human.

\begin{figure*}[t!]
\centering
    \includegraphics[width=\textwidth]{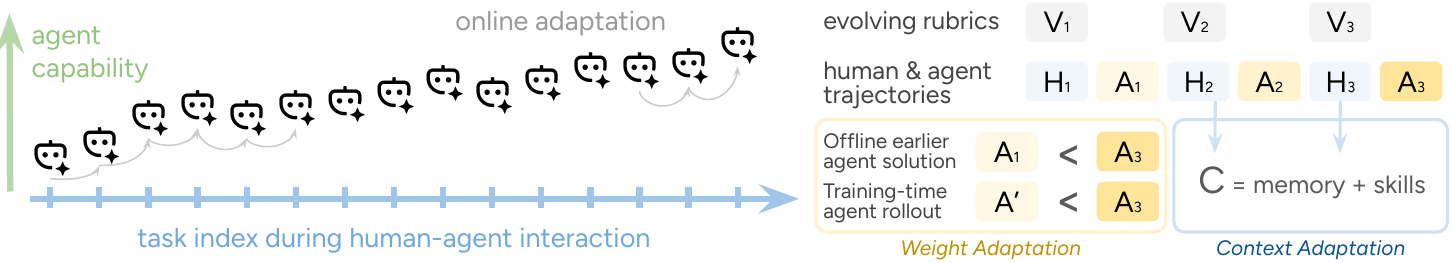}
\caption{Left: test-time agent adaptation paradigm, illustrated with an example agent capability measure. Right: Implementation overview of context- and weight-based adaptation.}
\label{fig:agent-adaptation}
\end{figure*}

Formally speaking, the adaptive agent processes a total of $T$ tasks one by one: $t = 1, \cdots, T$. 
The agent is instantiated with an unadapted model and empty context, denoted as $A^0$, and gets updated after every task. 
Specifically, after the human and agent $A^t$ co-finish the $t$-th task session, the agent $A^t$ uses this session data to upgrade into $A^{t+1}$ (\S\ref{sec:2.3:adapt-method}). Next, human interacts with the latest agent $A^{t+1}$ for the $t+1$-th task. 
As illustrated in \autoref{fig:agent-adaptation}, we expect the agent's capability to roughly improve throughout the process of solving the stream of tasks.

To evaluate test-time adaptive agent solo success on a given task $t$, we adopt the first final-step artifact $f^{t}_0$ produced solely by the agent in the first (index 0) task iteration $u^t_0$, and evaluate it against the final rubrics $V^{t}_J$ to report the average success rate of all verifier scores. 
Notably, the agent producing this first artifact has only received supervision signals from previous tasks, but is exposed to nothing but the task instruction about the current task.

As an ablated setup for this \textit{online} test-time agent adaptation, we also collect data from users interacting with non-adapting agents on the same $K$ tasks, denoted as the \textit{offline} data. We use this data to perform clean ablation studies on human interaction signals and agent consolidation strategies in later experiment sections.

\subsection{Agent Adaptation Methods}
\label{sec:2.3:adapt-method}
Given the agent policy $\pi^A_{\theta}(a | o, c)$, two adaptable components inside are the agent context $c$ and backbone LM weights $\theta$, which we denote as context- and weight-based adaptations.

\subsubsection{Context-Based Adaptation}
\label{sec:2.2.1:context-based}
Context-based adaptation transforms all human activities $\{\tau^H_{j}\}$ into verbalized textual information in two categories: (1) agent memory $M$ storing factual and preferential knowledge, and (2) agent skill library $K$ with procedural knowledge. At the $t$-th context adaptation, both inductions are powered by an LM conditioned on memory- and skill-specific instructions:
\begin{equation}
    \texttt{induce}_{LM_{\text memory}}(\{\tau^H_{j}\}, M^t) \rightarrow M^{t+1}, \qquad \texttt{induce}_{LM_{\text skill}}(\{\tau^H_{j}\}, K^t) \rightarrow K^{kt1}
\end{equation}
yielding an adapted policy $\pi^A_{\theta}(a | o, [M^t; K^t]) \rightarrow \pi^A_{\theta}(a | o, [M^{t+1}; K^{t+1}])$. 
Memory $M$ captures declarative, user- or domain-specific information such as ``the figure title should be bold-typed'' or ``the user prefers simple wording.''
Skill library $K$ encodes multi-step workflows into reusable procedure knowledge, e.g., ``to write a paper abstract: 1. write a motivating sentence, 2. introduce our method, 3. state numeric evidence, ...''. See \S\ref{app:lm-prompts} for our induction prompts. 

\subsubsection{Weight-Based Adaptation}
\label{sec:2.2.2:weight-based}
Weight-based adaptation integrates human activities into model parameters $\theta$. 
Unlike context-based adaptation that is bounded by what can be articulated as text and the model's ability to condition on it, weight-based approaches can potentially internalize extra patterns that are implicit and hard to verbalize, and directly train the model to produce different outputs.

Through preliminary exploration with various training algorithms (\S\ref{app:weight-explore}), we adopt Direct Preference Optimization (DPO) \citep{rafailov2023direct} and construct preference pairs by contrasting trajectories in the first and the last iterations $\tau_{0}$ and $\tau_{0:J}$, where $\tau_{0:j}$ means the concatenation of trajectories $[\tau_0, \tau_1, \cdots, \tau_j]$. 
To prevent agents from relying on human interaction to achieve the correct task solution, we consolidate $\tau_{0:j}$ into a shorter one-shot solution $\tau^{A*}_j$, by merging multiple intermediate file edits into a single edit action with multiple arguments, and skipping the observational (e.g., read) and trigger actions. In this way, we are training the agent to prefer generating the final high-quality solution in one pass, without conditioning on intermediate human actions. 
Achieving this would mean agents are capable of one-shot user-preferred solutions by maintaining proper user modeling, and do not rely on human supervision further.
Concretely,
\begin{equation}
    \mathcal{L}_{DPO}(\theta) = - \log \sigma \left(\beta \left( \log \frac{\pi_{\theta}(\tau^{A*}_{J} | q_j)}{\pi_{ref}(\tau^{A*}_{J} | q_j)} - \log \frac{\pi_{\theta}(\tau^{A*}_{j} | q_j)}{\pi_{ref}(\tau^{A*}_{j} | q_j} \right) \right), j = 0
\end{equation}
where
\begin{equation}
    \log \pi_{\theta}(\tau | q) = \sum_l log \pi_{\theta}(a_l | q, a_{<l}, s_{<l}; c)
\end{equation}
$a_l$ represents the action taken by the agent at step $l$, and $s_l$ represents the resulting environment state. $q_j$ is user query up to iteration $j$, in our setup, this is the initial task instruction when $j=0$.
$\sigma$ is the sigmoid function, $\beta$ is a hyper-parameter controlling how far $\pi_{\theta}$ can drift from $\pi_{ref}$. In implementation, we set the reference policy $\pi_{ref}$ to be the initial policy before training on the task session.
Although the policy does not explicitly condition on human activities, we do use the human activities to construct the supervision trajectory $\tau^{A*}_j$.
From each task session, we use the pair of the first and the last solutions as they perform the best empirically. 

Training on these constructed pairs provides a limited number of training data points, we therefore augment the training pool with pairs constructed by sampling agent solutions during training time. Particularly, we treat the agent solutions sampled at training time as the rejected trajectory, and pair it with the final solution in the corresponding collected task session as the chosen trajectory, to create additional pairs for DPO training.
See \S\ref{app:weight-explore} for more detailed comparisons across different pair construction strategies.

%% file: sections/3_expertise.tex
\section{Experiment: Adapting Agents to Human Expertise}
\label{sec:3:expertise}

In this section, we first introduce the experiment setup for test-time agent adaptation (\S\ref{sec:3.1:experiment-setup}), then demonstrate its effectiveness using both the context and weight update methods (\S\ref{sec:3.2:adapt-result}) with versatile human interaction signals (\S\ref{sec:3.3:human-interact-signals}). We further compare context and weight adaptation in their efficiency (\S\ref{sec:3.4:adapt-efficiency}) and learned expertise (\S\ref{sec:3.5:what-agent-learns}) to provide more insights.

\subsection{Tasks, Setup, and Evaluation}
\label{sec:3.1:experiment-setup}

\noindent \textbf{Tasks.} \quad We construct tasks for two scenarios where AI agents have been relatively widely adopted, namely (i) paper abstract writing, given the title and introduction section of the paper, and (ii) data visualization: given an instruction of data and intended visualization style, create an HTML figure to realize the visuals.\footnote{We are interested in domains (i) where certain amount of humans have greater, individualized expertise than an average AI agent, and (ii) are more open-ended to reflect the complex nature of real-world human digital activities.}
For each scenario, we manually constructed 20 tasks using the best and outstanding papers of NeurIPS, ICML, ICLR, *CL, EMNLP, and CHI conferences in 2025 to ensure content quality and diverse topic coverage.
Besides reporting test-time agent solo success rate on these 20 tasks, we further test how well agents can generalize the learned knowledge to unseen tasks. We similarly create another 30 held-out tasks for both writing and data visualization scenarios, with the best papers from the same conferences in 2023 and 2024, covering different topics from the 20 test-time tasks we collect data from (\S\ref{app:data-collection}).

\noindent \textbf{Setup} \quad 
For each task scenario, i.e., writing or data visualization, we recruit 5 human participants to interact with context-adaptive agents, and another 5 humans for weight-adaptive agents. To perform ablation studies, we recruit another 5 humans for non-adaptive agents, i.e., the offline setting. Each human interacts with the agent while completing 20 tasks within their assigned scenario.
In summary, we collect $2~\text{scenarios} \times 15~\text{users} \times 20~\text{tasks} = 600~\text{data points}$.

For paper abstract writing, we recruit CS PhD students in five universities who are second-year or above and have experience in academic writing on topics spanning ML, NLP, and HCI. For data visualization, we recruit from the Handshake talent network,\footnote{\url{https://joinhandshake.com/}} where each participant has substantial experience in data analysis from professional backgrounds in healthcare, business analysis, engineering, or research in math, psychology, or biology.

We initialize the agent with an empty context $c=\emptyset$ on top of the \texttt{Qwen3.6-35B-A3B} backbone LM hosted and trained with tinker API.\footnote{\url{https://tinker-docs.thinkingmachines.ai/tinker/}}
The context induction and evolving verifier are supported by \texttt{claude-sonnet-4-6}.
\footnote{We also tried contemporaneous open-source models in preliminary experiments, but they failed to produce reusable memory, skill, and verification criteria with desired quality.}

\noindent \textbf{Evaluation} \quad 
We test each test-time adapted agent in two folds. On the 20 tasks we collect interaction data on, we evaluate agent initial solutions $f^k_0$ for each task $k$ on its final evolved rubrics $V^k_J$, and calculate the average rubric scores throughout all tasks, to demonstrate its online adaptation ability.
Note that our agent is adapted toward individual users $e$. We report the average scores among all users in the context and weight groups, respectively.
To test if agent improvement generalizes across tasks, we also run the final adapted agent $A^K$ on the 30 held-out tasks without any human involvement or further adaptation, and similarly take its produced artifacts for evaluation. 
To evaluate how well an agent adapts toward its corresponding user $e$, we need to construct dedicated evaluation rubrics $V_e$ for these held-out tasks, as they are not similarly produced as the 20 test-time tasks we collect data on. 
Concretely, we ask the \texttt{claude-sonnet-4-6} LM to create rubrics for each task given its instruction, and ask it to cover (i) human-written rubrics: we ask user $e$ to write rubrics generally applicable to all tasks in their tasking scenario (i.e., abstract writing, data visualization),\footnote{Due to resource limitations to ask each of the 30 humans to annotate rubrics for all 30 tasks.} and (ii) summarized rubrics from the 20 tasks we collected from this user $\{V^k_e\}_{k=1}^{20}$.
For summarized rubrics (ii), we use \texttt{claude-sonnet-4-6} to extract the important criteria shared across 20 tasks we collected data from.
We ensure the generated held-out rubrics align with both the human-written and summarized rubrics, to ensure it satisfied this particular user's requirements.

In comparison, we run the \textit{baseline}, namely an unadapted agent $A^0$ solo on the same test-time and held-out tasks independently, and evaluate the produced artifacts with the same rubrics.

To get a sense of the upper-bound task performance the agent could possibly adapt to given this session data, we also take the last version of the artifact $f_J$ as the \textit{oracle} outcome from human-agent interaction. We similarly grade it with final rubrics $V_J$ and report the task average scores. 
Empirically, as we later show in \autoref{tab:overall-results}, this oracle score does not reach 100\% often, due to limitations in the agent capability or allowed human effort to realize all human requirements.

\subsection{Agents Can Efficiently Adapt to Human Expertise via Interaction}
\label{sec:3.2:adapt-result}

\input{artifacts/overall-results}

As shown in \autoref{tab:overall-results}, both adaptation approaches substantially improve agent solo task-solving performance within just 20 task sessions. On the tasks we collect human-agent interaction data on (test-time tasks), our context- and weight-adaptive agents --- evaluated on each task before the agent ever received human suggestions on it, using only adaptation accumulated from prior tasks --- improve their solo task-solving success rate relatively by 4.5\% and 4.5\% on writing tasks, 
and 12.9\% and 20.9\% on data visualization tasks
, respectively.
To demonstrate effective improvements on this relatively small pool of 100 data points per task under each adaptation scenario, we conducted paired t-tests between baseline and adapted agents on matched task instances to assess the statistical significance of these gains. Notably, every paired test reaches significance ($t > 2.0$), demonstrating that the adaptation is substantially effective even from a small number of examples.

To further test the cross-task generalization abilities of our adapted agents, we similarly evaluate baseline and adapted agents on held-out tasks unseen during data collection and agent adaptation.
As shown in \autoref{tab:overall-results} (held-out tasks), context- and weight-adaptive agents improve significantly over the baseline agent by  3.6\% and 5.7\% on writing and 2.4\% and 5.8\% on data visualization, relatively. 
These consistent improvements across task sets confirm that the adaptation generalized beyond the specific tasks during adaptation.

Comparing the two task domains, we observe larger gains in data visualization than in writing, plausibly because the backbone LM's writing ability is already more extensively developed through pretraining, leaving less headroom for adaptation to improve upon. 
When transferring to held-out tasks within the same domain, writing gains remain comparable to its test-time performance, whereas gains on unseen data visualization tasks shrink noticeably. This asymmetry suggests that data visualization tasks are heterogeneous, and exposure to a wider variety of tasks will likely facilitate adaptation more.

\subsection{Benefiting from Human Interaction Signals Beyond Feedback}
\label{sec:3.3:human-interact-signals}

\begin{wrapfigure}[11]{r}{0.5\textwidth}
\vspace{-3mm}    
\includegraphics[width=0.5\textwidth]{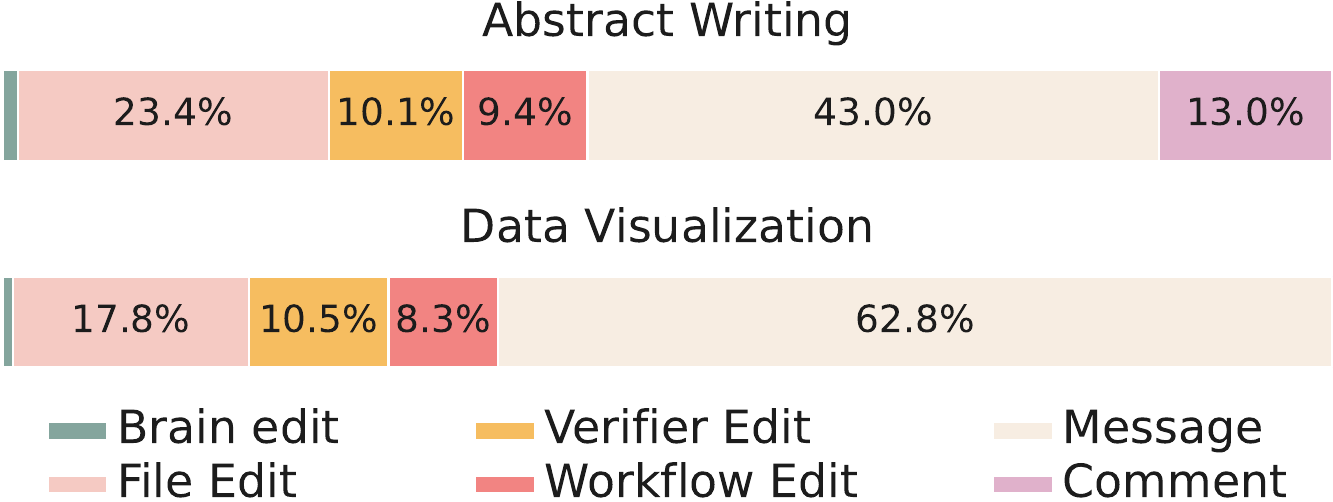}
\caption{Distribution of human actions throughout their interactions with the agents. \texttt{comment} was not supported for data visualization tasks.}
\label{fig:action-distribution}
\end{wrapfigure}

Much prior work reduced human feedback to textual messages sent back to the LM or agent \citep{christiano2017deep}. Yet humans naturally express feedback through a much richer repertoire of activity, most of which remain unexplored as a source of agent learning signal. Through our interface, beyond text \texttt{message} actions, experts can directly edit files, adjust the agent's plan, revise verification criteria, and leave inline comments; each forming a distinct channel through which expertise surfaces.
As shown in \autoref{fig:action-distribution}, humans draw on this full repertoire throughout their interaction with agents.

\begin{wraptable}[9]{r}{0.38\textwidth}
\vspace{-4mm}
\centering
\small
\resizebox{0.38\textwidth}{!}{
\begin{tabular}{l|cc}
\toprule
\multicolumn{1}{c|}{\bf Method} & {\bf Writing} & {\bf DataViz} \\
\midrule
{Baseline} & {80.8} & {66.4} \\
{All Actions} & {84.9} & {75.6} \\
{Msg Actions} & {81.9} & {68.0} \\
\bottomrule
\end{tabular}
}
\vspace{-1mm}
\caption{Comparing agent adaptation with all human actions and message actions alone.}
\label{tab:non-msg-actions}
\vspace{-1mm}
\end{wraptable}

To test whether these diverse human actions serve as effective adaptation signals, we compare agent adaptation results using (i) all human actions and (ii) message actions alone.  
We run this comparison on collected offline data using context adaptation, since it is difficult to isolate the effect of non-message actions in the online scenario, or under the weight adaptation approach.
(i) uses our default method for context adaptation, while (ii) restricts context induction input to message actions only.
As shown in \autoref{tab:non-msg-actions}, adding non-message actions yields 3.0\% and 7.6\% more gains in writing and data visualization tasks, showing the value of actions beyond text feedback.

\subsection{Weight Adaptation Has Greater Inference-Time Efficiency than Context Adaptation}
\label{sec:3.4:adapt-efficiency}

Beyond task performance, weight adaptation offers a significant efficiency advantage at inference time. 
On the input side, weight-adaptive agents require 62.6\% and 88.3\% fewer tokens than their context-adaptive counterpart on writing and data visualization tasks, showing a direct benefit of encoding expertise in model weights rather than an ever-growing context. 

\begin{figure*}[t!]
\centering
    \includegraphics[width=\textwidth]{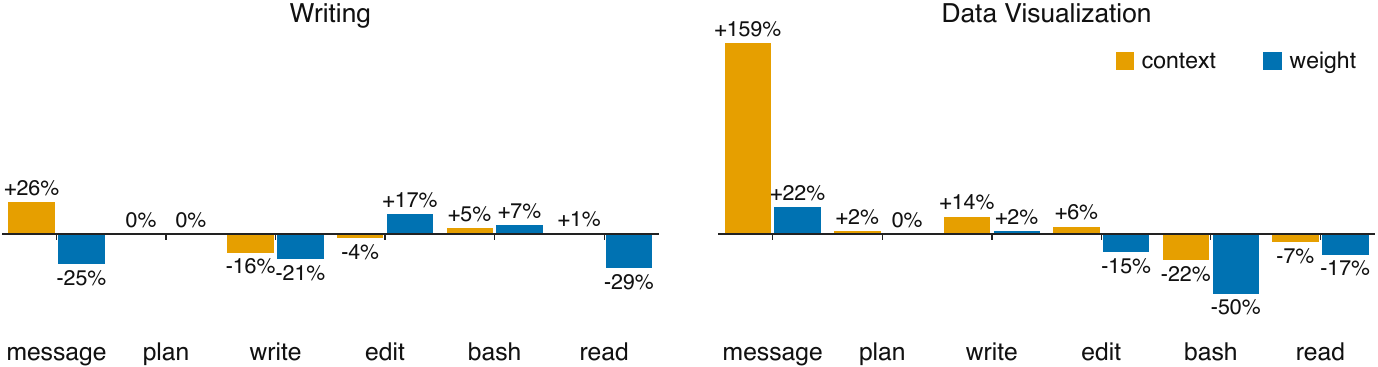}
\caption{Relative change in mean token count by action type, for context- and weight-adaptive agents compared to the unadapted baseline agent. Positive values indicate longer action content than baseline; negative values indicate shorter action content. For both abstract writing (left) and data visualization (right) tasks, the context-adaptive agent produces 143\% and 176\% longer messages than baseline and weight-adaptive agents.
}
\label{fig:token-count-by-actions}
\end{figure*}

This efficiency gain extends to the output side as well. Relative to task sessions from humans and unadapted baseline agents, weight-adaptive agents produce 9.1\% and 4.4\% fewer output tokens on writing and data visualization tasks, all while achieving better results. In contrast, context-adaptive agents move in the opposite direction, producing responses 20.1\% and 6.1\% longer on the same tasks, suggesting that context adaptation inherits some of the verbosity of the contexts it conditions on.
In \autoref{fig:token-count-by-actions}, we investigate the deeper reason by analyzing the action content length by different action types, and find context-adaptive agents produce 143--176\% and 69.9--95.7\% longer \texttt{message}s, while rarely differ in other task execution actions, compared to the baseline and weight-adaptive agent, respectively.

\subsection{What Agents Learned, and Failed to Learn, from Humans}
\label{sec:3.5:what-agent-learns}

We analyze the passed and failed rubrics of our adapted agents against baseline and oracle, to examine what our agents learned, and failed to learn, through adaptation. Particularly, we compare test-time solutions produced by (i) the baseline agent, (ii) our adapted agents, and (iii) human-agent interaction (i.e., oracle). 
We measure \textit{Agent Gains} by finding rubrics fulfilled by adapted agents (ii) yet failed by the baseline (i), and \textit{Gaps to Oracle} with rubrics failed by adapted agents (ii) that are achieved in oracle solutions (iii). 
For both writing and data visualization tasks, we categorize these rubrics by themes, as where humans expressed their expertise (\S\ref{app:agent-adaptation-details}).

For writing tasks, humans expressed most of their expertise in \textit{discourse precision and compression} (19.6\%), \textit{surface form and venue conventions} (19.1\%), and \textit{context specificity} (17.6\%), reflecting practical knowledge of venue-implicit conventions that experts frequently correct agents toward, along with a persistent emphasis on precise, concise writing.

Among themes, \textit{surface form} and \textit{terminology and naming} are easiest to absorb, with 70.4--73.7\% of expressed expertise successfully integrated, likely because they are easy to specify explicitly and straightforward for agents to adhere to. 
\textit{Context specificity} and \textit{problem framing}, in contrast, prove harder to learn, with only 41.9--43.4\% of expressed expertise captured, suggesting these categories demand more implicit, situational judgment that is harder to transfer in a handful of examples.
Comparing the two adaptation methods, context and weight acquire largely comparable knowledge, with similar proportions across themes, contrary to prior works that assign distinct functions to context and weights \citep{tiwari2026learning}.

For data visualization, humans express expertise predominantly around textual annotations on \textit{titles, legends, and labels} (38.5\%), along with visual aspects such as \textit{color} (15.5\%) and \textit{spatial clarity} (12.4\%). 
Notably, humans show particular expertise in what to highlight in a figure and how to highlight it (\textit{color and visual encoding}), where both adaptation methods exhibit a substantial gap from the human oracle, leaving 60.1\% and 50.0\% of expertise uncaptured.
In contrast, implementation-oriented aspects are markedly easier for agents to integrate: 75.2\% of \textit{data fidelity} requirements and 94.4\% of \textit{HTML conventions} requirements are fulfilled.

\begin{figure*}[t!]
\centering
    \includegraphics[width=\textwidth]{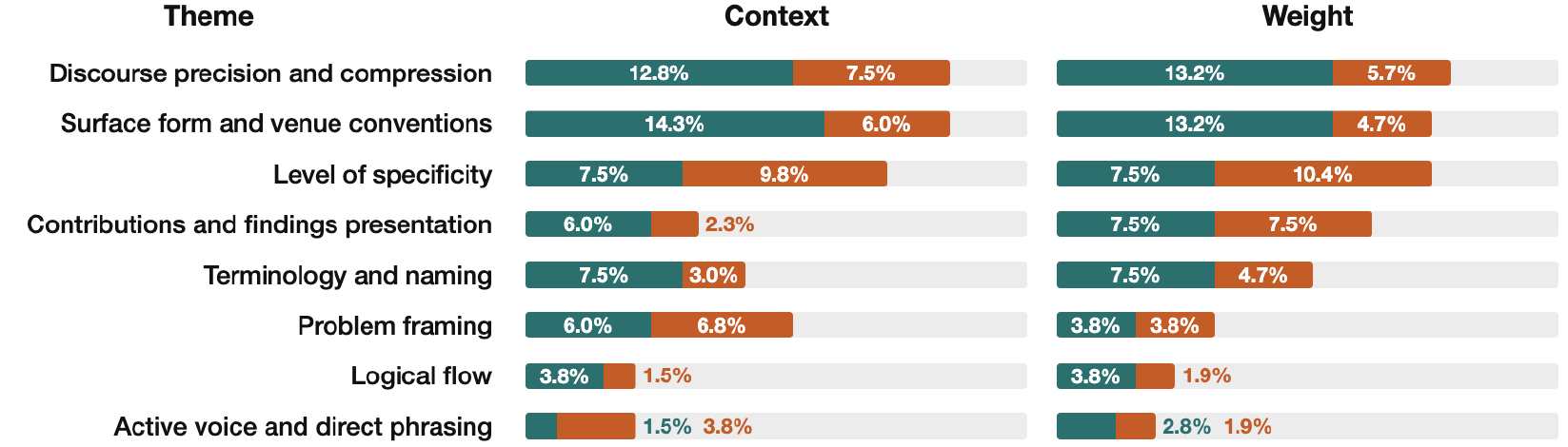}
    
    \vspace{4mm}
    
    \includegraphics[width=\textwidth]{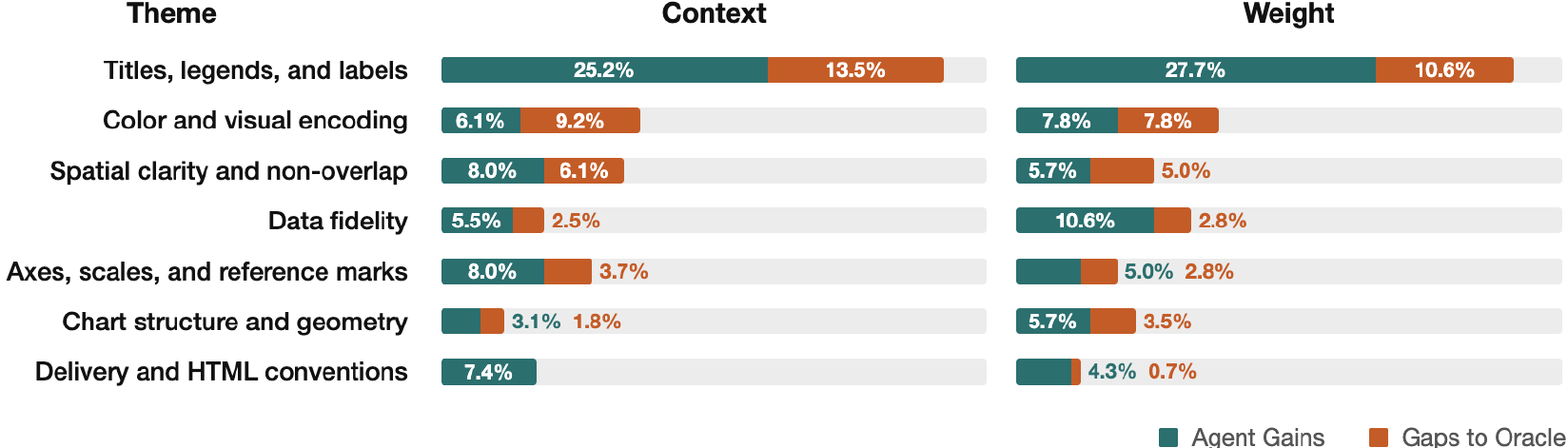}
\caption{Expertise absorbed by agents' contexts and weights during adaptation (\textit{Agent Gains}) and remaining from the human oracles (\textit{Gap to Oracle}), on writing (top) and data visualization (bottom) tasks. See \S\ref{app:agent-adaptation-details} for detailed descriptions of each expertise theme.}
\label{fig:expertise-adapt-gap}
\end{figure*}

%% file: artifacts/overall-results.tex
\begin{table}[t!]
\centering
\small
\resizebox{0.95\linewidth}{!}{
\begin{tabular}{ll|ccc|cc}
\toprule 
\multicolumn{1}{c}{\multirow{2}{*}{\bf Scenario}} & \multicolumn{1}{c}{\multirow{2}{*}{\bf Category}} & \multicolumn{3}{c}{\bf Test-Time Tasks} & \multicolumn{2}{c}{\bf Held-Out Tasks} \\
{} & {} & {Baseline} & {Adaptive} & {Oracle} & {Baseline} & {Adaptive} \\
\midrule
\multirow{2}{*}{Context} & {Abstract Writing} & {81.7} & {85.4$^{*}~~$} & {89.5$^{***}$} & {90.2} & {93.5$^{***}$} \\
{} & {Data Visualization} & {77.0} & {86.9$^{***}$} & {95.0$^{***}$} & {89.9} & {92.1$^{*}~~$} \\
\midrule
\multirow{2}{*}{Weight} & {Abstract Writing} & {82.8} & {86.5$^{*}~~$} & {91.1$^{***}$} & {88.0} & {93.0$^{***}$} \\
{} & {Data Visualization} & {69.0} & {81.1$^{***}$} & {93.8$^{***}$} & {86.2} & {91.2$^{**}~$} \\
\bottomrule
\end{tabular}
}
\caption{Comparing agent solo success rates (SR) on paper abstract writing tasks and data visualization tasks, with test-time online adaptation via context (top) and weight (bottom). 
\textit{Test-Time Tasks} report agents' performance on the 20 tasks we collected data on; \textit{Held-Out Tasks} report agents' performance on 30 held-out tasks of the same categories. 
We performed significance testing on \textit{adaptive} and \textit{oracle} methods against the \textit{baseline}, specifically two-sided paired t-tests. 
Notably, all adaptation results are statistically significantly better than the baseline, i.e., |t-statistic| $> 2.0$. We use $^{*}$, $^{**}$, $^{***}$ to indicate results with $p<0.05$, $p<0.01$, and $p<0.001$.
} 
\label{tab:overall-results}
\end{table}

%% file: sections/4_rubrics.tex
\section{Eliciting Human Expertise with Scalable Rubric Evolvement}
\label{sec:4:rubric-validation}

In this section, we switch focus to the evolving verifier module in our agent suite, and demonstrate its potential to serve as a scalable rubric annotation tool for open-ended tasks (\S\ref{sec:4.1:rubric-comparison}). We further dissect the rubrics into compositions of shared community expertise and individual strategies and preferences (\S\ref{sec:4.2:rubric-composition}), and show that our adapted agents learned both (\S\ref{sec:4.3:agent-learns-both}).

\subsection{Rubrics Derived from Human-Agent Interaction Capture More Failures than LLM-Only and Human-Only Rubrics}
\label{sec:4.1:rubric-comparison}

Tasks beyond mathematical and programming problems have long been treated as ``non-verifiable,'' owing to the difficulty of annotating a rubric that is both comprehensive and accurate enough for evaluation.
How to properly evaluate such open-ended tasks remains a long-standing problem: recent work has largely converged on LLM-as-a-judge \citep{zheng2023judging} scored against rubrics, produced either by LLMs or humans. Yet each approach suffers from a distinct set of shortcomings.

For one, LLM-created rubrics often miss aspects that human experts care about and are prone to self-bias, favoring LLM-produced solutions over ones a human evaluator would prefer \citep{deutsch2022limitations,shankar2024validates}.
However, human-only annotation is no easier, as human judgment is naturally inconsistent on open-ended tasks, and producing high-quality rubrics demands substantial effort to instruct, motivate, and supervise annotators \citep{xu2024theagentcompany,vidgen2026apex}. Worse, annotators are typically asked to write rubrics from task instructions alone, ungrounded in actual execution, making them prone to miss details that only surface once the task is executed \citep{xu2024theagentcompany}.
To this end, we design our evolving rubric module to capture human expertise directly from their interaction with the agent, while relieving the manual annotation burden via automatic LLM-based rubric generation for scalability. 

\begin{figure*}[t!]
\centering
    \includegraphics[width=0.99\textwidth]{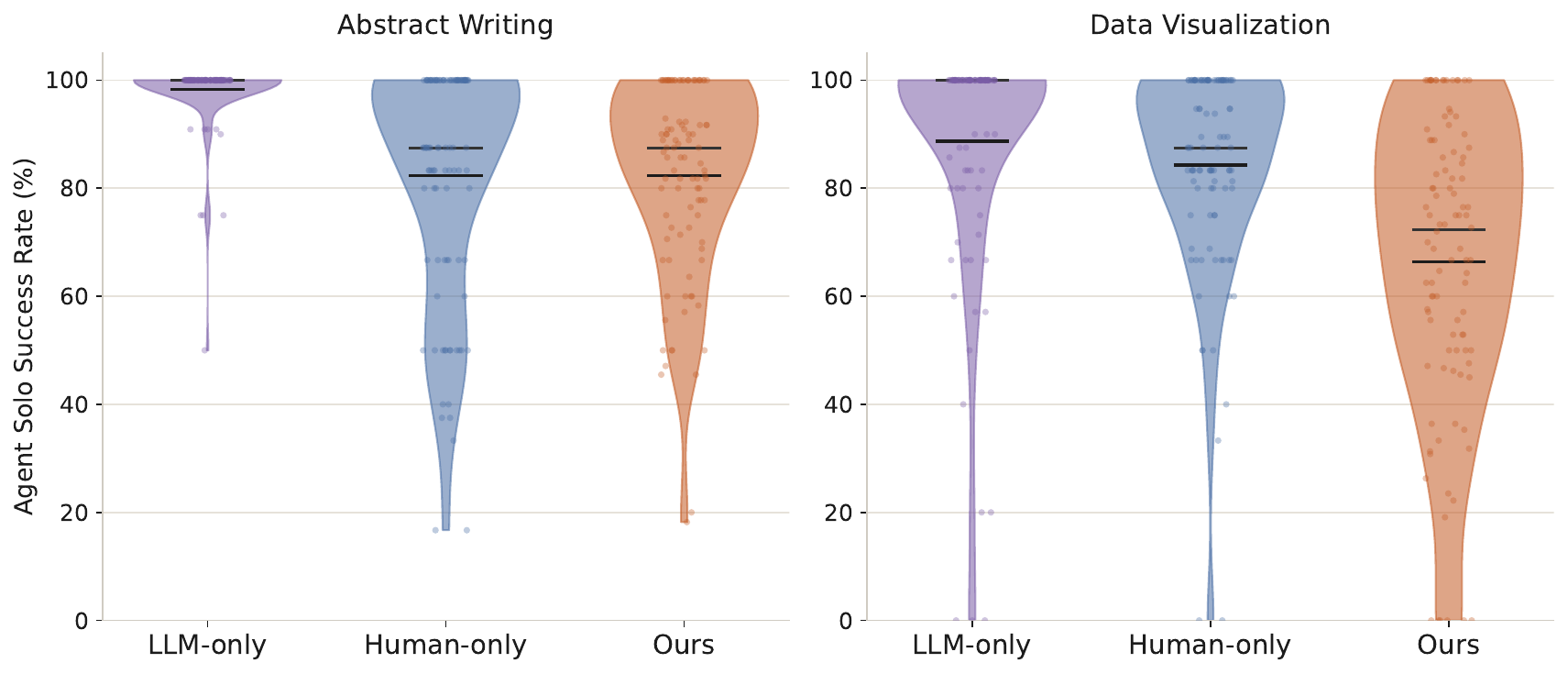}
\caption{LLM-only rubrics fail to distinguish low-quality agent solutions (left: writing, right: data visualization), whereas our rubrics capture comparable overlooked aspects to the human-written rubrics for writing, even substantially more overlooked aspects for data visualization.}
\label{fig:rubrics-validation}
\end{figure*}

To show this empirically, we collect two baseline rubric sets for comparison: (i) LLM-produced rubrics given the task instruction, and (ii) human-written rubrics given the task scenario.\footnote{Given limited human resources to annotate rubrics across 100 tasks, we have humans annotate one rubric set that can be applied to all tasks in their assigned category (i.e., writing or data visualization) generally.}
We then evaluate each rubric's ability to capture failures in agents' solo solutions, namely the agent's initial, unassisted attempt at each task session, prior to any human interaction. Since these solo solutions are later revised through numerous rounds of human iterations within the same session, we take this as evidence that they are imperfect and should score accordingly low against a truly comprehensive rubric.
Concretely, we evaluate agent solo solutions on each of the three sets of rubrics, and show their scores in \autoref{fig:rubrics-validation}. 

First, scores against LLM-only rubrics saturate at an average of 98.3\% and 88.7\% at writing and data visualization tasks, respectively. Our evolved rubrics, by contrast, yield substantially lower average scores of 82.3\% and 66.4\%, suggesting they capture more failures the baseline rubrics miss.
Compared to human-only rubrics, our evolved rubrics capture a comparable share of failures on writing tasks, but 17.9\% more on data visualization tasks. Manually analyzing the human-only rubrics, we conjecture this gap stems mainly from a difference in annotator motivation and expertise, as we hired senior CS PhD students for the writing tasks, but general human workers for data visualization. Thus, the LM-based rubric evolution may have more room to improve over the human baseline in data visualization, as general workers may be less well trained and motivated to annotate evaluation rubrics. 
Find more validation tests in \S\ref{app:rubrics-validation}.

\subsection{Rubrics Capture Shared Community Expertise and Individual Expertise}
\label{sec:4.2:rubric-composition}

Even when interacting with the same initial agent on the same set of tasks, different humans arrive at different solutions and different rubrics. This reflects a basic property of open-ended tasks such as the writing and data visualization we investigated: they admit no single, universal notion of correctness. In fact, countless solutions can solve such a task well, that is, score highly against their corresponding evaluation criteria, depending on what the particular expert prefers. 
Despite this interpersonal variance, most fields still maintain community guidelines broadly agreed upon by their practitioner, forming a notion of relative correctness.
We hypothesize that the evolved rubrics by our interface capture both components:
(i) shared community guidelines, such as grounding every written claim in concrete numerical evidence; and (ii) individual strategies and preferences, such as one expert favoring enumeration while another avoids it. 

\begin{figure*}[t!]
\centering
    \includegraphics[width=0.49\textwidth]{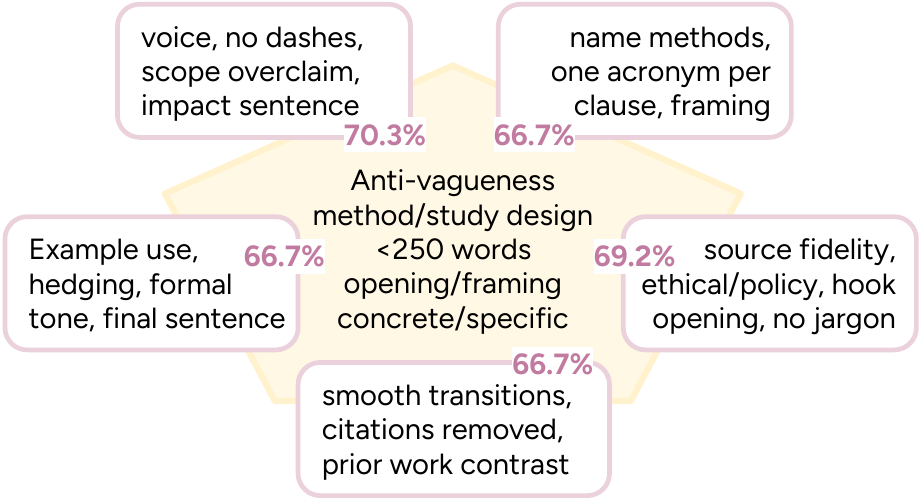}
    \hfill
    \includegraphics[width=0.49\textwidth]{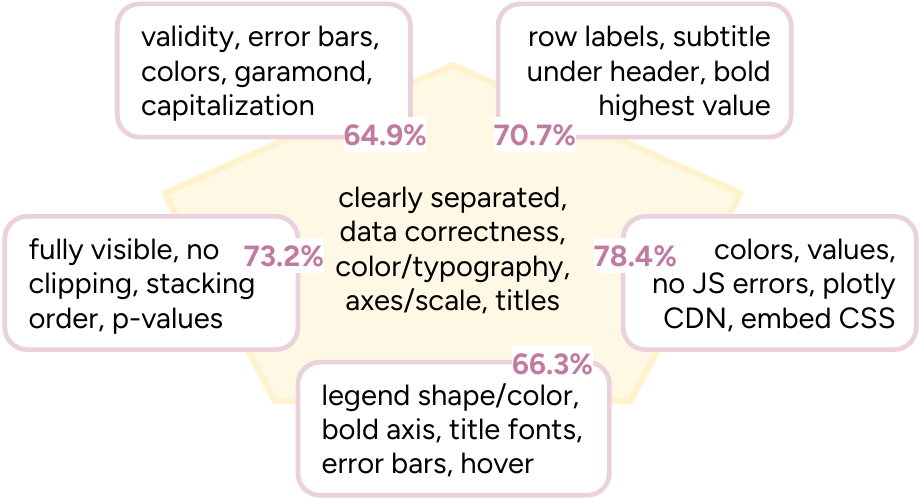}
\caption{Shared and personalized rubrics on abstract writing (left) and data visualization (right) tasks. Each intersection value indicates the percentage of that user's rubrics classified as shared, out of all rubrics derived from their interaction.}
\label{fig:expertise-venn}
\end{figure*}

To test this hypothesis, we take each user's evolved rubrics and, for every criterion, check whether it is also covered by other users' rubrics on the same task. A criterion covered by over 60\% of other rubrics is categorized as a shared guideline; otherwise, we treat it as personalized.
Across all writing and data visualization tasks, we find an average of 67.9\% and 70.7\% of user rubrics fall under shared community guidelines, with the remainder reflecting personalized expertise and preferences.
\autoref{fig:expertise-venn} illustrates these shared criteria alongside exemplar personalized ones. In both tasks, we observe users placing emphasis on different aspects. E.g., for visualizations, some users prioritize implementation details (e.g., Plotly CDN, CSS styling), while others emphasize statistical rigor, such as error bars and significance testing.

%% file: sections/5_community.tex
\subsection{Agent Learns Both Shared and Personalized Expertise}
\label{sec:4.3:agent-learns-both}

\begin{wraptable}[10]{r}{0.55\textwidth}
\vspace{-4mm}
\centering
\small
\resizebox{0.54\textwidth}{!}{
\begin{tabular}{l|cccc}
\toprule
\multirow{2}{*}{\bf Method} & \multicolumn{2}{c}{\bf Writing: $\Delta$SR\%} & \multicolumn{2}{c}{\bf DataViz: $\Delta$SR\%} \\
{} & {Shared} & {Personal} & {Shared} & {Personal} \\
\midrule
{Context} & {0.3} & {6.2} & {6.1} & {19.6} \\
{Weight} & {1.9} & {9.4} & {8.8} & {13.8} \\
\bottomrule
\end{tabular}
}
\vspace{-1mm}
\caption{Agents adapted to individual experts improve on both personalized criteria and shared community criteria, relative to the baseline agent, on writing (left) and data visualization (right) tasks.}
\label{tab:agent-learn-both-rubrics}
\vspace{-1mm}
\end{wraptable}

A natural question follows: through this efficient adaptation, does the agent only learn individual expertise, or does it also improve on community-agreed standards more broadly?
For each agent adapted to a specific expert, we separately measure its success rates on shared and personalized criteria, and compare both against the baseline agent's success rates.
As shown in \autoref{tab:agent-learn-both-rubrics}, beyond the significant gains on personalized criteria, agents also improve by up to 8.8\% on shared criteria for the two tasks, signaling that this adaptation yields genuine gains in fundamental capability, beyond just individual fit.

\subsection{Exploration: Deriving Community Expertise from Individual Adaptations}
\label{sec:4.4:consolidation}

Although each agent is adapted towards a particular user, we've shown that an average of 69.3\% of users' expertise reflects shared community guidelines. We thus ask: can individually adapted agents be consolidated to improve performance across users broadly?
In this section, we explore initial strategies for merging multiple individualized agents into a single joint agent, in both context and weight channels, and demonstrate the open challenge for effective merging.

\noindent \textbf{Context Merging} \quad
We use an LLM (\texttt{claude-sonnet-4-6}) to merge the contexts derived from different users' interaction data into a single, shared piece of content. We then augment this shared content onto the baseline agent, and evaluate it on held-out tasks.

\noindent \textbf{Weight Merging} \quad
We explored two approaches to consolidate model weights derived from various users. 
The first, \textit{Weight-T}rain, pools all users' data to train a single model from scratch.  
The second, \textit{Weight-M}erge, avoids retraining by directly merging the individually trained weights, concretely by averaging the LoRA adapter parameters \citep{wortsman2022model} and combine it with the base model for direct inference.

\begin{wraptable}[9]{r}{0.70\textwidth}
\vspace{-2mm}
\centering
\small
\resizebox{0.69\textwidth}{!}{
\begin{tabular}{l|ccc|ccc}
\toprule
\multirow{2}{*}{\bf Method} & \multicolumn{3}{c|}{\bf Writing: $\Delta$SR\%} & \multicolumn{3}{c}{\bf DataViz: $\Delta$SR\%} \\
{} & {Shared} & {Personal} & {Overall} & {Shared} & {Personal} & {Overall} \\
\midrule
{Context} & {-4.4} & {1.3} & {$~$-2.7} & {2.9} & {-0.5} & {+1.1} \\
{Weight-T} & {-4.7} & {3.5} & {$~$-2.3} & {3.6} & {$~$2.7} & {+3.3} \\
{Weight-M} & {$~$0.0} & {3.2} & {+1.0} & {3.6} & {$~$1.8} & {+3.1} \\
\bottomrule
\end{tabular}
}
\caption{Exploring different approaches to consolidate individualized agents into one.}
\label{tab:community-expertise}
\vspace{-1mm}
\end{wraptable}

We conduct all merging with data collected from the offline scenario, since conducting a clean Weight-Train experiment is difficult with adapting agents in the online scenario.
As shown in \autoref{tab:community-expertise}, all merging methods improve over the baseline on data visualization tasks, but yield smaller gains on writing tasks. Breaking down shared and personal rubrics, we find that on writing tasks, all merging methods satisfy personal criteria better than the baseline, yet exhibit varying performance in preserving shared criteria. Yet for data visualization tasks, variance mainly comes from meeting the personal criteria.
This may be because baseline agents already perform strongly on writing, leaving less room for improvement that can generalize across users. Data visualization is comparably more open-ended, thus stressing models on divergent human expertise, whose effective integration remains an open challenge \citep{park2024rlhf,chakraborty2024maxminrlhf}.

Another potential factor is the offline nature of our adaptation: compared with online adaptation, where the agent receives human feedback targeted at its current state, offline adaptation uses potentially outdated human feedback thus leads to smaller gains \citep{tang2024understanding}.
Together, these findings highlight the challenge of effectively bringing versatile individual expertise into a single agent and motivate future work for expertise consolidation.

%% file: sections/discussions.tex
\section{Related Work}

\paragraph{Test-Time Agent Adaptation}
Due to the persistent gap between training data and the versatility of downstream usage, adapting agents at test time has become increasingly popular for meeting ad-hoc downstream requirements \citep{gao2026a}. A large portion of this work has focused on adapting agent context. On one hand, existing studies induce procedural \citep{wang2025agent}, factual \citep{packer2023memgpt}, and preferential knowledge into agent memory augmented at inference time. On the other hand, for further quality guarantees and efficiency, programmatic skills have been designed for robust knowledge reuse \citep{wang2025inducing,zheng2025skillweaver,xia2026skillrl}.
Context-based adaptation dominates this space largely because it is easier to interpret and implement, while still yielding observable improvements from just a handful of task sessions. Partly for this reason, updating the weights of the agent backbone LM remains far less explored, constrained by limited available data as well as the difficulty of implementation and inspection \citep{lin2025continual}. As a result, despite the simplicity of context adaptation, little is known about how it compares to weight-based test-time training on a per-instance basis. 
Recent studies introduce test-time training (TTT, \citealt{sun2020test}) to realize this online model weight update, leveraging weak supervision signals from experimented tasks \citep{yuksekgonul2026learning}, yet still require substantial amounts of training to be effective.
Our work challenges the presumed inferiority and data requirements of weight updates, demonstrating its feasibility on agentic systems beyond the backbone LMs.

\paragraph{Learning from Human Signals}
A long-standing line of work trains LMs using human binary preferences or NL feedback to improve instruction-following abilities \citep{christiano2017deep,chiang2024chatbot,buening2026aligning}. In agent training specifically \citep{alvinn1988pomerleau}, many studies focus on training agents to imitate human demonstrations \citep{ross2011reduction,wang2025opencua,shaikh2025aligning}. 
However, human activity typically involves UI-level interaction, which differs drastically from the programmatic approaches that LM-based agents favor \citep{wang2025how}, making raw human demonstrations a poor fit for agent learning. To address this mismatch, some works instead have humans provide feedback on agent-driven solutions and translate that feedback into training signals \citep{wang2026openclaw}. While effective, these approaches typically require substantial amounts of data, which is often infeasible in real-world scenarios, overlooking many alternative action-taking signals from human activities.
In contrast, our work develops an efficient adaptation strategy that meaningfully improves agent performance within just tens of task sessions.

\paragraph{Evaluating Agents on ``Non-Verifiable'' Tasks}
LM agents experienced a first surge of development on software engineering tasks, where execution-based, unit test-style verifiers \citep{chen2021evaluating} have become the standard and robust way to evaluate agent performance \citep{jimenez2024swe}.
However, as we move from these easily verifiable tasks to more open-ended tasks such as creative writing or visual creation, two main challenges emerge. First, we need to find alternative ways for evaluation: human evaluation is often unstable and hard-to-scale \citep{patwardhan2026gdpval}; some studies have attempted to use LMs to directly produce a score \citep{liu2023geval}, or more robustly, provide a set of binary judgments against a rubric \citep{aggarwal2026gym,shao2026collabskill}.
Constructing high-quality rubrics is difficult, since for open-ended tasks, it becomes unclear what ``task success'' means, as there exist countless ``correct'' ways to solve these tasks \citep{jiang2026artificial}.
While most works leverage LLMs or hire human annotators to construct rubrics, our experiments show that they are incomplete due to LM knowledge limitations and human lacking grounding in task execution. To address this, we propose an LLM-automated, human-in-the-loop rubric evolution strategy, to create high-quality rubrics for both model training and agent evaluation purposes.

\paragraph{Training Agentic Language Models}
Different from training a text-producing LM, agentic LMs need to consume observation-style inputs such as serialized web page content or UI-style visuals, as well as produce structured responses containing executable actions and explanatory thoughts. 
Beyond the proliferation of works around the low-agentic regime such as question answering or chat-style tasks \citep{lin2025continual}, recent work has investigated strategies to synthesize a great volume of data \citep{song2026agent} to train agents for common purposes such as software engineering \citep{pan2025training,jain2025regym} and web navigation tasks \citep{ou2024synatra,shen2025thinking,murty2024bagel,zhou2024archer}. 
Others turn to human-sourced data such as recorded human computer-use activities \citep{he2026efficient,wang2026opencua}, yet still suggest one of the most critical recipes is constructing scalable, quality data for training.
However, this is not always feasible in practice, as it may be expensive, time- or cost-wise, to prepare numerous data for a specific downstream application. This raises an under-explored question: can we effectively update model weights by exploiting a limited number of examples?
In this work, we directly tackle this problem and show that it's possible.

\section{Conclusion}

In this work, we introduce an adaptive agent framework that performs test-time adaptation via context and weight updates using human-agent interaction data. Throughout experiments with 20 human experts, our agents effectively adapt to individual task success within merely 20 interactions. Our evolving verifier module automatically incorporates human interaction signals and produces more comprehensive evaluation rubrics for realistic, open-ended tasks. Both adapted agents and evolved rubrics capture two layers of expertise: community expertise shared across users, and individual expertise specific to a user.

Overall, our results point to a concrete design principle for future AI agents: learn continuously from test-time interaction, and adapt to the individual humans they serve.
We hope this work motivates research into more advanced test-time agent adaptation techniques, as well as systematic documentation and benchmarking of community- and individual-level expertise.


\section*{Acknowledgments}
Zora is supported by the Google PhD Fellowship. This work was partially supported by the National Science Foundation under award number 2543679. We thank Tinker Research Grants for supporting the weight-time agent adaptation experiments.
We thank people in the Language Technologies Institute at CMU and the Stanford NLP Group for their helpful discussions and feedback throughout the project.


%% file: appendix/2_problem_method.tex
\section{Data Collection}
\label{app:data-collection}

For 20 tasks we collect human-agent interaction data on, we present their venues and topics in \autoref{tab:title-20-tasks}.
In \autoref{tab:title-heldout-tasks}, we also list out the source of 30 held-out tasks, demonstrating their varied distributions from the 20 tasks above, and their validity in serving as our generalization test bed.

\begin{table}[htbp]
\centering
\resizebox{\textwidth}{!}{%
\begin{tabular}{cl}
\toprule
\multicolumn{1}{c}{\bf Venue} & \multicolumn{1}{c}{\bf Title} \\
\midrule
{NeurIPS} & {Artificial Hivemind: The Open-Ended Homogeneity of Language Models (and Beyond)} \\
{NeurIPS} & {Gated Attention for Large Language Models: Non-linearity, Sparsity, and Attention-Sink-Free} \\
{NeurIPS} & {1000 Layer Networks for Self-Supervised RL: Scaling Depth Can Enable New Goal-Reaching Capabilities} \\
{NeurIPS} & {On the Generalization Properties of Diffusion Models} \\
{ICLR} & {Safety Alignment Should Be Made More Than Just a Few Tokens Deep} \\
{ICLR} & {Learning Dynamics of LLM Finetuning} \\
{ICLR} & {AlphaEdit: Null-Space Constrained Knowledge Editing for Language Models} \\
{ICML} & {CollabLLM: From Passive Responders to Active Collaborators} \\
{ICML} & {Train for the Worst, Plan for the Best: Understanding Token Ordering in Masked Diffusions} \\
{ICML} & {Roll the dice \& look before you leap: Going beyond the creative limits of next-token prediction} \\
{ICML} & {Conformal Prediction as Bayesian Quadrature} \\
{ICML} & {Score Matching With Missing Data} \\
{ICML} & {The Value of Prediction in Identifying the Worst-Off} \\
{ACL} & {INFINI-GRAM MINI: Exact n-gram Search at the Internet Scale with FM-Index} \\
{ACL} & {Mind the Value-Action Gap: Do LLMs Act in Alignment with Their Values?} \\
{ACL} & {LINGGYM: How Far Are LLMs from Thinking Like Field Linguists?} \\
{ACL} & {Generative or Discriminative? Revisiting Text Classification in the Era of Transformers} \\
{ACL} & {Measuring Chain of Thought Faithfulness by Unlearning Reasoning Steps} \\
{CHI} & {Synthetic Human Memories: AI-Edited Images and Videos Can Implant False Memories and Distort Recollection} \\
{CHI} & {Creative Writers' Attitudes on Writing as Training Data for Large Language Models} \\
\bottomrule
\end{tabular}
}
\caption{Titles and venues of the 20 tasks we collected human-agent interaction data on.}
\label{tab:title-20-tasks}
\end{table}

\begin{table}[htbp]
\centering
\resizebox{\textwidth}{!}{%
\begin{tabular}{cl}
\toprule
\multicolumn{1}{c}{\bf Venue} & \multicolumn{1}{c}{\bf Title} \\
\midrule
{ICLR} & {Transformers are Inherently Succinct} \\
{ICLR} & {LLMs Get Lost In Multi-Turn Conversation} \\
{CHI} & {Towards Fluent Interaction with Cyber-Physical Architecture} \\
{CHI} & {``I Don’t Think RAI Applies to My Model'' - Engaging Non-champions with Sticky Stories for Responsible AI Work} \\
{CHI} & {iTagPDF: Towards Finally Automating PDF Accessibility} \\
{NeurIPS} & {Visual Autoregressive Modeling: Scalable Image Generation via Next-Scale Prediction} \\
{NeurIPS} & {Not All Tokens Are What You Need for Pretraining} \\
{NeurIPS} & \makecell[l]{The PRISM Alignment Dataset: What Participatory, Representative and Individualised Human \\Feedback Reveals About the Subjective and Multicultural Alignment of Large Language Models} \\
{ICLR} & {Generalization in Diffusion Models Arises from Geometry-Adaptive Harmonic Representations} \\
{ICLR} & {Never Train from Scratch: Fair Comparison of Long-Sequence Models Requires Data-Driven Priors} \\
{ICLR} & {Protein Discovery with Discrete Walk-Jump Sampling} \\
{ICML} & {Position: Considerations for Differentially Private Learning with Large-Scale Public Pretraining} \\
{ICML} & {Debating with More Persuasive LLMs Leads to More Truthful Answers} \\
{ICML} & {Genie: Generative Interactive Environments} \\
{ACL} & {Mission: Impossible Language Models} \\
{ACL} & {Why are Sensitive Functions Hard for Transformers?} \\
{ACL} & {Aya Model: An Instruction Finetuned Open-Access Multilingual Language Model} \\
{ACL} & {How Johnny Can Persuade LLMs to Jailbreak Them} \\
{EMNLP} & {Pretraining Data Detection for Large Language Models: A Divergence-based Calibration Method} \\
{EMNLP} & {An Image Speaks a Thousand Words, but Can Everyone Listen? On Image Transcreation for Cultural Relevance} \\
{EMNLP} & {Toward Robust Speech Representation Learning for Thousands of Languages} \\
{EMNLP} & {KidLM: Advancing Language Models for Children} \\
{CHI} & {Constrained Highlighting in a Document Reader can Improve Reading Comprehension} \\
{NeurIPS} & {Scaling Data-Constrained Language Models} \\
{NeurIPS} & {Direct Preference Optimization: Your Language Model is Secretly a Reward Model} \\
{ICLR} & {Emergence of Maps in the Memories of Blind Navigation Agents} \\
{ICLR} & {Universal Few-shot Learning of Dense Prediction Tasks with Visual Token Matching} \\
{ACL} & {Do Androids Laugh at Electric Sheep? Humor ``Understanding'' Benchmarks from The New Yorker Caption Contest} \\
{ACL} & {Marked Personas: Using Natural Language Prompts to Measure Stereotypes in Language Models} \\
{ACL} & {Weaker Than You Think: A Critical Look at Weakly Supervised Learning} \\
\bottomrule
\end{tabular}
}
\caption{Titles and venues of the held-out tasks we tested agent generalization abilities on.}
\label{tab:title-heldout-tasks}
\end{table}

%% file: appendix/method-ablation.tex
\section{Exploration on Weight-Based Adaptation Approaches}
\label{app:weight-explore}
We conducted a broad exploration of weight adaptation approaches and selected DPO as the approach used in the main experiments (\S\ref{sec:2.3:adapt-method}) based on its superior performance.

\subsection{Alternative Algorithms}
Besides DPO, we also experimented with on-policy distillation (OPD) and REINFORCE training, as formulated below.

\noindent \textbf{On-Policy Distillation (OPD)} \quad
applies a token-level KL divergence target from a teacher policy $\pi_{teacher}$ on the current agent policy $\pi_{\theta}$. We adopt on-policy self-distillation (OPSD) \citep{zhao2026selfdistilled}, where the teacher adopts the same parameterization $\theta$ and the teacher ``privilege'' comes from the additional context $\{\tau^A_{j'}, \tau^H_{j'}\}_{j'=j}^J$ it has access to, namely later iterations of agent and human activities in the session. Formally speaking, compared to the student policy $\pi_{\theta}(\cdot | q)$, the teacher policy is $\pi_{teacher}(\cdot | q) = \pi_{\theta}(\cdot | q, \{\tau^A_{j'}, \tau^H_{j'}\}_{j'=j}^J)$. 
The learning loss can be represented as:
\begin{align}
    \mathcal{L}_{OPD}(\theta) = - \sum_l \sum_{t \in a_{j,l}} \pi_{teacher}(a_{j,l} | q, a_{j,<l}, s_{j,<l}) \log \pi_{\theta}(a_{j,l} | q, a_{j,<l}, s_{j,<l}) \\
    = - \sum_l \sum_{t \in a_{j,l}} \pi_{\theta}(a_{j,l} | q, a_{j, <l}, s_{j, <l} ; \{\tau^A_{j'}, \tau^H_{j'}\}_{j'=j}^J ) \log \pi_{\theta}(a_{j,l} | q, a_{j,<l}, s_{j,<l})
\end{align}

\paragraph{REINFORCE} uses the scalar reward $r_j = \sum(\{r_j\})$ of the $j+1$-th agent trajectory $\tau^A_j$, optionally with quantified human activities (e.g., number of actions $r_j = |\tau^H_j|$), as a policy gradient signal to reinforce high-reward trajectories. We represent the collective reward as $R_j = \sum(\{r_j\}) - \lambda |\tau^H_j|$, where $\lambda$ is the importance factor of human activities. The overall REINFORCE loss can be formalized as:
\begin{equation}
    \mathcal{L}_{REINFORCE}(\theta) = - (R_j - b) \sum_l \log \pi_{\theta}(a_{j,l} | q_j, a_{j,<l}, s_{j,<l})
\end{equation}
where $b$ is the baseline reward to reduce variance, we implement $b$ as the reward from last iteration $b = R_{j-1}$.

\paragraph{Performance Analysis}
In preliminary experiments, we find both alternatives perform worse than DPO. 
For REINFORCE, we find that agents rarely explore high-scoring solutions according to the evolved rubrics via human feedback, yet constantly synthesize solutions that stuck around a 0.3--0.4 success score range out of 1. We conjecture that agents have limited exposure to expert-style solutions in their training data, therefore hard to explore such solutions and impossible to hill-climb even given the rubrics without additional execution hints.
In comparison, OPD has the advantage of providing privileged information to the teacher policy to hint at the student exploration. Nonetheless, at least in our data-efficient regime, we find that directly training the agent policy to prefer to generate the oracle solution (i.e., DPO) is more effective than learning the representation of it (i.e., OPD)

\subsection{DPO Data Formulation}
For DPO training, we compared multiple strategies of constructing pairwise data. Specifically, we experimented with (i) first-last: pairing the first and the last solution, (ii) enumerate: pairing every earlier and later solution possible, (iii) min-gap-k: pairing earlier-later solutions that are at least $k$ iterations apart. Empirically, we find (i) performs the best, so we adopted this setup in both offline and on-policy pair construction in the main approach in \S\ref{sec:2.2:adapt-paradigm}.

%% file: appendix/3_agent-adaptation.tex
\section{Human Expertise Learned During Agent Adaptation}
\label{app:agent-adaptation-details}

In \autoref{tab:expertise-desc-writing} and \autoref{tab:expertise-desc-dataviz}, we provide descriptions of the expertise categories shown in \autoref{fig:expertise-adapt-gap}, on writing and data visualization tasks, respectively.

\begin{table}[htbp]
\centering
\resizebox{\textwidth}{!}{%
\begin{tabular}{ll}
\toprule
\multicolumn{1}{c}{\bf Category} & \multicolumn{1}{c}{\bf Exemplar Task} \\
\midrule
{Surface form and venue conventions} & \makecell[l]{Mechanical presentation rules including word count limits (150-250 words), \\single-paragraph layout, removal of section headings, punctuation \\conventions (no em-dashes), and informal register avoidance.} \\
\midrule
{Active voice and direct phrasing} & \makecell[l]{Preference for first-person active constructions ('we find', 'we show', 'we \\introduce') over passive voice, direct statements over vague constructions, \\and accessible terminology that prioritizes clarity for the target audience. } \\
\midrule
{Terminology and naming} & \makecell[l]{Handling of acronyms (expansion on first mention with parenthetical \\abbreviations), consistent technical naming, jargon definition or avoidance, \\precise use of domain terms, and elimination of ambiguous references or \\undefined terminology.} \\
{Problem framing} & \makecell[l]{Covers how the abstract opens with the research problem or gap, establishes \\motivation and context.} \\
\midrule
{Level of specificity} & \makecell[l]{Requirements for grounded detail including specific numbers, dataset names, \\model scales, measurement units, sample sizes, named examples of domains \\or tasks, quantitative results, and concrete experimental parameters rather \\than abstract descriptions.} \\
\midrule
{Contributions and findings presentation} & \makecell[l]{What the work claims to deliver or demonstrate, including listed \\contributions as distinct points, headline findings with concrete support, key \\insights explicitly highlighted, theoretical contributions stated, and \\implications or takeaways for the field.} \\
\midrule
{Discourse precision and compression} & \makecell[l]{Combines tight, precise wording that bans hedges, vague qualifiers, and \\unsupported superlatives with discourse compression that enforces one idea \\per sentence, eliminates redundancy and repetition, and condenses filler \\without sacrificing clarity.} \\
\midrule
{Logical flow} & \makecell[l]{distinguish between different types of claims or comparisons, separate \\distinct ideas into their own sentences or clauses. maintains logical \\progression from problem to method to results with explicit transitions and \\narrative coherence across sentences.} \\
\bottomrule
\end{tabular}
}
\caption{Description of observed human expertise categories in writing tasks.}
\label{tab:expertise-desc-writing}
\end{table}

\begin{table}[htbp]
\centering
\resizebox{\textwidth}{!}{%
\begin{tabular}{ll}
\toprule
\multicolumn{1}{c}{\bf Category} & \multicolumn{1}{c}{\bf Exemplar Task} \\
\midrule
{Delivery and HTML conventions} & \makecell[l]{Mechanical packaging of the deliverable: valid self-contained HTML, required library/\\CDN tags, presence of expected canvas or chart elements, and browser-renderable output \\without errors. Differs from Chart structure and geometry, which covers how the chart \\is composed once it renders.} \\
\midrule
{Chart structure and geometry} & \makecell[l]{Recipe of the visualization itself: chart type, orientation, number and arrangement of\\ panels/series/groups, mark shape and sizing (bars, cells, heatmaps), grouping layout, \\and viewport dimensions. Differs from Delivery (file/HTML validity), Axes (numeric \\scales), and Data fidelity (exact encoded values).} \\
\midrule
{Data fidelity} & \makecell[l]{Exact match of encoded values to the specified data: cell contents, bar heights/lengths, \\error-bar extents, series/row/column counts, and correctly placed best-value markers. \\Differs from Color and visual encoding (how values are colored) and from Chart structure \\(layout recipe, not the numbers).} \\
\midrule
{Axes, scales, and reference marks} & \makecell[l]{Numeric scaffolding of the plot: axis ranges, tick intervals and order, linear vs log scaling, \\presence or absence of gridlines, and dashed/threshold reference lines at specified positions. \\Differs from Titles, legends, and labels (axis title/tick text wording) and from Data fidelity \\(the data values plotted on those axes).} \\
\midrule
{Color and visual encoding} & \makecell[l]{Mapping of values and groups onto color and related visual channels: semantic color \\mappings (e.g. green=desirable), colormaps and opacity, series/group color families, \\highlight fills, and contrast against the background. Differs from Data fidelity (whether \\values are correct) and from Titles, legends, and labels (legend text explaining colors).} \\
\midrule
{Titles, legends, and labels} & \makecell[l]{Textual and annotative content: chart/section titles, axis and series names, legend content \\and placement, abbreviations/naming conventions, captions, headers, plus overlay marks \\and affordances (significance brackets, stars/asterisks, arrows, dotted guides, hover/\\tooltips). Differs from Spatial clarity (whether text/marks overlap or clip) and from Color \\(the color mapping itself, not the legend wording).} \\
\midrule
{Spatial clarity and non-overlap} & \makecell[l]{Layout hygiene so elements stay visible and distinct: label placement that avoids occlusion, \\no clipping of bars/error bars/tooltips, cell borders, margins/padding, and legend bounds. \\Differs from Titles, legends, and labels (what text/marks say or which ones exist) and from \\Chart structure (panel/viewport recipe rather than overlap/clipping).} \\
\bottomrule
\end{tabular}
}
\caption{Description of observed human expertise categories in data visualization tasks.}
\label{tab:expertise-desc-dataviz}
\end{table}

%% file: appendix/4_rubrics.tex
\section{Rubrics Validation}
\label{app:rubrics-validation}

Besides effectively capturing failures in imperfect task solutions (\S\ref{sec:4:rubric-validation}), we show our rubrics are as effective in revealing task success as other solutions, further validating the quality of our evolved rubrics.

\begin{figure*}[t!]
\centering
    \includegraphics[width=0.99\textwidth]{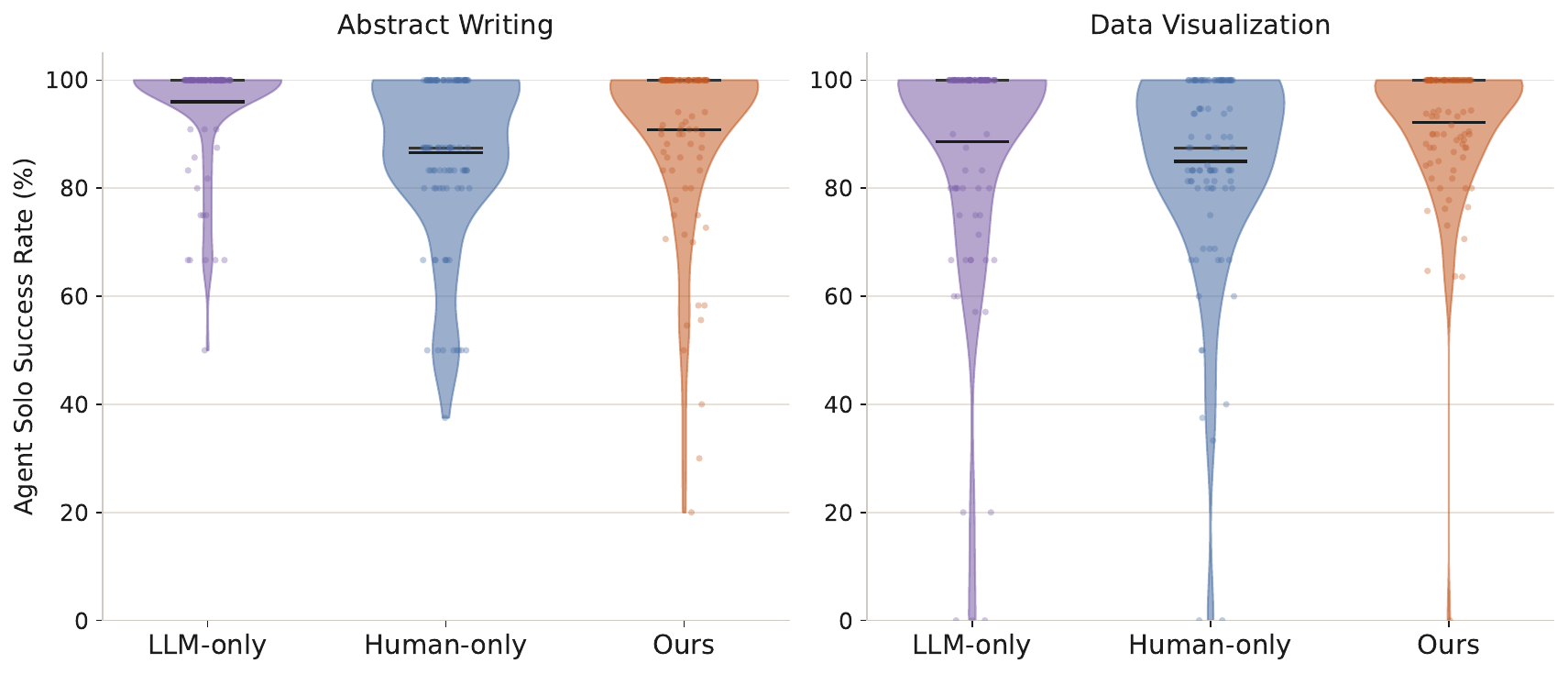}
\caption{Our evolved rubrics are as effective in measuring task success as LLM and human-produced rubrics.}
\label{fig:rubrics-validation-oracle}
\end{figure*}

In \autoref{fig:rubrics-validation-oracle}, we evaluate the oracle solutions using LLM-only, human-only, and our rubrics, all of which are effective in signaling task success by producing scores around high scores. 
Taken together with results in \autoref{fig:rubrics-validation}, our rubrics fail imperfect agent solutions due to issues in the solution, rather than issues in the evaluation criteria.

%% file: appendix/lm-prompts.tex
\section{Prompts for LM-Supported Modules}
\label{app:lm-prompts}

\begin{tcolorbox}[colback=gray!10, colframe=gray!60, coltitle=black, title=Prompt for Memory Induction for Context-Adaptive Agents, fonttitle=\bfseries]
Merge new induction entries into the existing memory file and produce a refined, cross-session version.

You receive (i) the original memory file and (ii) new entries derived from a new session.
Merge (ii) into (i): keep durable prior preferences, add genuinely new ones, and lightly deduplicate.
Output line count should stay about the same as the original memory (i). Do not add excessive new entries.

Keep:
- Specific user preferences that apply to certain contexts
- Recurring styling or workflow habits (e.g. larger fonts, no gridlines, compact layout) across tasks
- General facts about how the user works

Remove or merge only when necessary:
- Nonsensical or contradictory entries
- Highly task-specific details unlikely to help elsewhere (e.g. a color for one named column/bar)
- Duplicate lines that say the same thing in different words.
- Make each line concise but do not over-compress away useful detail.

Output rules:
- One entry per line. Plain text only (NO markdown headers like \# or \#\#).
- Prefix each line with "Fact:" or "Preference:".
- Do not include reasoning or a thinking process.

Reply with:
Title: <short topic name>
- Fact: <item>
- Preference: <item>
\end{tcolorbox}

\begin{tcolorbox}[colback=gray!10, colframe=gray!60, coltitle=black, title=Prompt for Skill Induction for Context-Adaptive Agents, fonttitle=\bfseries]
From the task and numbered log, describe the workflow the agent used: ordered steps, generalized (no long paths).

Your primary job is to capture what THIS session did—especially techniques, fixes, and steps visible in the Log. The existing skill file—if any—is background only.

When an existing skill file is provided:
- FIRST update the workflow using concrete steps from this session's Log (mandatory when the log is non-empty).
- Add or revise steps for anything new in this session (e.g. chart tweaks, file edits, verification, user-requested changes).
- Adopt the useful parts from original steps only when still accurate; merge duplicates.
- Do not return the existing skill unchanged if the log shows new agent work.
Output the full updated skill (not a diff). Generalize: no long paths, file paths, or raw code.

Reply with:
Title: <short task name>
1. <step>
2. <step>
...
If nothing fits: NONE
\end{tcolorbox}

\begin{tcolorbox}[colback=gray!10, colframe=gray!60, coltitle=black, title=Prompt for Grading Solutions Against Rubrics, fonttitle=\bfseries]
You are an automated checker for completed task output files.
Given verifier criteria and current output files, decide whether each criterion is satisfied.
Reply with ONLY a JSON object of this exact shape: {"results":[{"pass":true},{"pass":false},...]}
{Task Instruction}

Verifier criteria (in order):
{Criteria 1}
{Criteria 2}
...

Output files and contents: {File Content}
\end{tcolorbox}

\begin{tcolorbox}[colback=gray!10, colframe=gray!60, coltitle=black, title=Prompt for Summarizing Test-Time Task Rubrics, fonttitle=\bfseries]
You distill grading rubrics for {task category} tasks.

You are given:
(A) High-quality, task-specific verifiers written/edited by a user across many past tasks.
(B) A fixed list of human-written general guidelines that every visualization should satisfy.

Your job: produce a JOINT GENERAL verifier list for this expertise overall.

Rules:
- Capture recurring themes from the user's history (layout, labels, legends, annotations,
  spacing/overlap, axes/scales, colors, HTML validity, browser renderability,
  chart-type-specific checks when they recur).
- Phrase items as general, reusable criteria (not tied to one past task's numbers/labels).
- Do NOT include criteria whose only job is checking that plotted numeric values match
  an instruction (those are task-specific). Prefer structural/visual quality checks.
- MUST cover every human guideline in (B). Either keep them (possibly tightened) or rewrite
  them so the same requirement is clearly present. Do not drop a human guideline.
- Prefer the user's concrete, falsifiable style from (A) over vague wording.
- Deduplicate near-duplicates; keep one clear criterion per idea.
- Typically 10–20 items. Prefer fewer clear lines over a long union.

Reply with JSON only (no markdown fences, no commentary):
{
  "general\_verifiers": ["...", "..."],
  "human\_coverage": [
    {"human": "<exact human guideline>", "covered\_by": "<matching general\_verifiers item>"}
  ]
}
Every human guideline must appear exactly once in human\_coverage.
\end{tcolorbox}

\begin{tcolorbox}[colback=gray!10, colframe=gray!60, coltitle=black, title=Prompt for Determining Shared Rubrics, fonttitle=\bfseries]
You compare grading rubrics for the same abstract-writing task.

You are given:
(A) Source rubrics — criteria from one user's verifier list for this task.
(B) Target rubrics — criteria from another user's verifier list for the SAME task.

For EACH source rubric, decide whether it is COVERED by the target list:
- covered=1 if at least one target rubric checks the same requirement
  (exact match, paraphrase, or a target rubric that subsumes the source check).
- covered=0 if no target rubric addresses that requirement.

Rules:
- Match on meaning, not wording. "under 250 words" covers "Word count < 250".
- A single target rubric may cover multiple source rubrics, and vice versa.
- Partial overlap that still enforces the core constraint counts as covered.
- Do not invent coverage from vague thematic similarity alone.
- Return one judgment per source rubric, in the same order as (A).

Reply with JSON only (no markdown fences, no commentary):
{
  "coverage": [
    {
      "source\_index": 0,
      "source": "<exact source rubric text>",
      "covered": 1,
      "covered\_by": "<exact target rubric text, or empty string if covered=0>",
      "note": "<brief reason>"
    }
  ]
}
Every source rubric must appear exactly once, in order, with covered as 0 or 1.
\end{tcolorbox}

%% file: main.bib
@article{xu2024theagentcompany,
  title={Theagentcompany: benchmarking llm agents on consequential real world tasks},
  author={Xu, Frank F and Song, Yufan and Li, Boxuan and Tang, Yuxuan and Jain, Kritanjali and Bao, Mengxue and Wang, Zora Z and Zhou, Xuhui and Guo, Zhitong and Cao, Murong and others},
  journal={arXiv preprint arXiv:2412.14161},
  year={2024}
}

@inproceedings{patwardhan2026gdpval,
  title={Gdpval: Evaluating ai model performance on real-world economically valuable tasks},
  author={Patwardhan, Tejal and Dias, Rachel and Proehl, Elizabeth and Kim, Grace and Wang, Michele and Watkins, Olivia and Fishman, Simon and Aljubeh, Marwan and Thacker, Phoebe and Fauconnet, Laurance and others},
  booktitle={International Conference on Learning Representations},
  volume={2026},
  pages={24005--24040},
  year={2026}
}

@inproceedings{wang2025agent,
title={Agent Workflow Memory},
author={Zora Zhiruo Wang and Jiayuan Mao and Daniel Fried and Graham Neubig},
booktitle={Forty-second International Conference on Machine Learning},
year={2025},
url={https://openreview.net/forum?id=NTAhi2JEEE}
}

@article{wang2025opencua,
  title={Opencua: Open foundations for computer-use agents},
  author={Wang, Xinyuan and Wang, Bowen and Lu, Dunjie and Yang, Junlin and Xie, Tianbao and Wang, Junli and Deng, Jiaqi and Guo, Xiaole and Xu, Yiheng and Wu, Chen Henry and others},
  journal={arXiv preprint arXiv:2508.09123},
  year={2025}
}

@inproceedings{wang2025inducing,
title={Inducing Programmatic Skills for Agentic Tasks},
author={Zora Zhiruo Wang and Apurva Gandhi and Graham Neubig and Daniel Fried},
booktitle={Second Conference on Language Modeling},
year={2025},
url={https://openreview.net/forum?id=lsAY6fWsog}
}

@inproceedings{jain2025regym,
title={R2E-Gym: Procedural Environment Generation and Hybrid Verifiers for Scaling Open-Weights {SWE} Agents},
author={Naman Jain and Jaskirat Singh and Manish Shetty and Tianjun Zhang and Liang Zheng and Koushik Sen and Ion Stoica},
booktitle={Second Conference on Language Modeling},
year={2025},
url={https://openreview.net/forum?id=7evvwwdo3z}
}

@misc{skills2025,
  title={Agent skills},
  author={Anthropic},
  howpublished={\url{https://agentskills.io/home}},
  year={2025}
}

@article{Gobet1998ExpertMA,
  title={Expert memory: a comparison of four theories.},
  author={Fernand R. Gobet},
  journal={Cognition},
  year={1998},
  volume={66 2},
  pages={115-52},
  url={https://api.semanticscholar.org/CorpusID:449552}
}

@inproceedings{alvinn1988pomerleau,
author = {Pomerleau, Dean A.},
title = {ALVINN: an autonomous land vehicle in a neural network},
year = {1988},
publisher = {MIT Press},
address = {Cambridge, MA, USA},
booktitle = {Proceedings of the 2nd International Conference on Neural Information Processing Systems},
pages = {305–313},
numpages = {9},
series = {NIPS'88}
}

@article{wang2025how,
  title={How Do AI Agents Do Human Work? Comparing AI and Human Workflows Across Diverse Occupations},
  author={Zora Zhiruo Wang and Yijia Shao and Omar Shaikh and Daniel Fried and Graham Neubig and Diyi Yang},
  journal={ArXiv},
  year={2025},
  volume={abs/2510.22780},
  url={https://api.semanticscholar.org/CorpusID:282388757}
}

@article{christiano2017deep,
  title={Deep reinforcement learning from human preferences},
  author={Christiano, Paul F and Leike, Jan and Brown, Tom and Martic, Miljan and Legg, Shane and Amodei, Dario},
  journal={Advances in neural information processing systems},
  volume={30},
  year={2017}
}

@article{vidgen2026apex,
  title={APEX-agents},
  author={Vidgen, Bertie and Mann, Austin and Fennelly, Abby and Stanly, John Wright and Rothman, Lucas and Burstein, Marco and Benchek, Julien and Ostrofsky, David and Ravichandran, Anirudh and Sur, Debnil and others},
  journal={arXiv preprint arXiv:2601.14242},
  year={2026}
}

@article{aggarwal2026gym,
  title={Gym-anything: Turn any software into an agent environment},
  author={Aggarwal, Pranjal and Neubig, Graham and Welleck, Sean},
  journal={arXiv preprint arXiv:2604.06126},
  year={2026}
}

@article{lin2025continual,
  title={Continual learning via sparse memory finetuning},
  author={Lin, Jessy and Zettlemoyer, Luke and Ghosh, Gargi and Yih, Wen-Tau and Markosyan, Aram and Berges, Vincent-Pierre and O{\u{g}}uz, Barlas},
  journal={arXiv preprint arXiv:2510.15103},
  year={2025}
}

@article{zheng2025skillweaver,
  title={Skillweaver: Web agents can self-improve by discovering and honing skills},
  author={Zheng, Boyuan and Fatemi, Michael Y and Jin, Xiaolong and Wang, Zora Zhiruo and Gandhi, Apurva and Song, Yueqi and Gu, Yu and Srinivasa, Jayanth and Liu, Gaowen and Neubig, Graham and others},
  journal={arXiv preprint arXiv:2504.07079},
  year={2025}
}

@article{xia2026skillrl,
  title={Skillrl: Evolving agents via recursive skill-augmented reinforcement learning},
  author={Xia, Peng and Chen, Jianwen and Wang, Hanyang and Liu, Jiaqi and Zeng, Kaide and Wang, Yu and Han, Siwei and Zhou, Yiyang and Zhao, Xujiang and Chen, Haifeng and others},
  journal={arXiv preprint arXiv:2602.08234},
  year={2026}
}

@article{gao2026a,
title={A Survey of Self-Evolving Agents: What, When, How, and Where to Evolve on the Path to Artificial Super Intelligence},
author={Huanang Gao and Jiayi Geng and Wenyue Hua and Mengkang Hu and Xinzhe Juan and Hongzhang Liu and Shilong Liu and Jiahao Qiu and Xuan Qi and Qihan Ren and Yiran Wu and Hongru WANG and Han Xiao and Yuhang Zhou and Shaokun Zhang and Jiayi Zhang and Jinyu Xiang and Yixiong Fang and Qiwen Zhao and Dongrui Liu and Cheng Qian and Zhenhailong Wang and Minda Hu and Huazheng Wang and Qingyun Wu and Heng Ji and Mengdi Wang},
journal={Transactions on Machine Learning Research},
issn={2835-8856},
year={2026},
url={https://openreview.net/forum?id=CTr3bovS5F},
note={Survey Certification}
}

@article{packer2023memgpt,
  title={Memgpt: Towards llms as operating systems},
  author={Packer, Charles and Wooders, Sarah and Lin, Kevin and Fang, Vivian and Patil, Shishir G and Stoica, Ion and Gonzalez, Joseph E},
  journal={arXiv preprint arXiv:2310.08560},
  year={2023}
}

@inproceedings{pan2025training,
title={Training Software Engineering Agents and Verifiers with {SWE}-Gym},
author={Jiayi Pan and Xingyao Wang and Graham Neubig and Navdeep Jaitly and Heng Ji and Alane Suhr and Yizhe Zhang},
booktitle={Forty-second International Conference on Machine Learning},
year={2025},
url={https://openreview.net/forum?id=Cq1BNvHx74}
}

@inproceedings{song2026agent,
title={Agent Data Protocol: Unifying Datasets for Diverse, Effective Fine-tuning of {LLM} Agents},
author={Yueqi Song and Ketan Ramaneti and Zaid Sheikh and Ziru Chen and Boyu Gou and Tianbao Xie and Yiheng Xu and Danyang Zhang and Apurva Gandhi and Fan Yang and Joseph Liu and Tianyue Ou and Zhihao Yuan and Frank F. Xu and Shuyan Zhou and Xingyao Wang and Xiang Yue and Tao Yu and Huan Sun and Yu Su and Graham Neubig},
booktitle={The Fourteenth International Conference on Learning Representations},
year={2026},
url={https://openreview.net/forum?id=tG6301ORHd}
}

@article{chen2021evaluating,
  title={Evaluating large language models trained on code},
  author={Chen, Mark and Tworek, Jerry and Jun, Heewoo and Yuan, Qiming and Pinto, Henrique Ponde De Oliveira and Kaplan, Jared and Edwards, Harri and Burda, Yuri and Joseph, Nicholas and Brockman, Greg and others},
  journal={arXiv preprint arXiv:2107.03374},
  year={2021}
}

@inproceedings{jimenez2024swe,
  title={Swe-bench: Can language models resolve real-world github issues?},
  author={Jimenez, Carlos E and Yang, John and Wettig, Alexander and Yao, Shunyu and Pei, Kexin and Press, Ofir and Narasimhan, Karthik},
  booktitle={International Conference on Learning Representations},
  volume={2024},
  pages={54107--54157},
  year={2024}
}

@article{shao2026collabskill,
  title={CollabSkill: Evaluating Human-Agent Collaboration On Real-World Tasks},
  author={Shao, Yijia and Wang, Zora Zhiruo and Ahuja, Neel and Wang, Yicheng and Liu, Bowen and Yang, Diyi},
  journal={arXiv preprint arXiv:2606.09833},
  year={2026}
}

@inproceedings{jiang2026artificial,
title={Artificial Hivemind: The Open-Ended Homogeneity of Language Models (and Beyond)},
author={Liwei Jiang and Yuanjun Chai and Margaret Li and Mickel Liu and Raymond Fok and Nouha Dziri and Yulia Tsvetkov and Maarten Sap and Yejin Choi},
booktitle={The Thirty-ninth Annual Conference on Neural Information Processing Systems Datasets and Benchmarks Track},
year={2026},
url={https://openreview.net/forum?id=saDOrrnNTz}
}

@article{zhou2024archer,
  title={Archer: Training language model agents via hierarchical multi-turn rl},
  author={Zhou, Yifei and Zanette, Andrea and Pan, Jiayi and Levine, Sergey and Kumar, Aviral},
  journal={arXiv preprint arXiv:2402.19446},
  year={2024}
}

@inproceedings{ou2024synatra,
title={Synatra: Turning Indirect Knowledge into Direct Demonstrations for Digital Agents at Scale},
author={Tianyue Ou and Frank F. Xu and Aman Madaan and Jiarui Liu and Robert Lo and Abishek Sridhar and Sudipta Sengupta and Dan Roth and Graham Neubig and Shuyan Zhou},
booktitle={The Thirty-eighth Annual Conference on Neural Information Processing Systems},
year={2024},
url={https://openreview.net/forum?id=KjNEzWRIqn}
}

@article{murty2024bagel,
  title={Bagel: Bootstrapping agents by guiding exploration with language},
  author={Murty, Shikhar and Manning, Christopher and Shaw, Peter and Joshi, Mandar and Lee, Kenton},
  journal={arXiv preprint arXiv:2403.08140},
  year={2024}
}

@article{wang2026opencua,
  title={Opencua: Open foundations for computer-use agents},
  author={Wang, Xinyuan and Wang, Bowen and Lu, Dunjie and Yang, Junlin and Xie, Tianbao and Wang, Junli and Deng, Jiaqi and Guo, Xiaole and Xu, Yiheng and Wu, Chen and others},
  journal={Advances in Neural Information Processing Systems},
  volume={38},
  pages={139756--139806},
  year={2026}
}

@inproceedings{he2026efficient,
title={Efficient Agent Training for Computer Use},
author={Yanheng He and Jiahe Jin and Pengfei Liu},
booktitle={The Fourteenth International Conference on Learning Representations},
year={2026},
url={https://openreview.net/forum?id=cDuA6ZNvCl}
}

@article{shen2025thinking,
  title={Thinking vs. doing: Agents that reason by scaling test-time interaction},
  author={Shen, Junhong and Bai, Hao and Zhang, Lunjun and Zhou, Yifei and Setlur, Amrith and Tong, Shengbang and Caples, Diego and Jiang, Nan and Zhang, Tong and Talwalkar, Ameet and others},
  journal={arXiv preprint arXiv:2506.07976},
  year={2025}
}

@article{chiang2024chatbot,
  title={Chatbot arena: An open platform for evaluating llms by human preference},
  author={Chiang, Wei-Lin and Zheng, Lianmin and Sheng, Ying and Angelopoulos, Anastasios Nikolas and Li, Tianle and Li, Dacheng and Zhang, Hao and Zhu, Banghua and Jordan, Michael and Gonzalez, Joseph E and others},
  journal={arXiv preprint arXiv:2403.04132},
  year={2024}
}

@inproceedings{shaikh2025aligning,
  title={Aligning language models with demonstrated feedback},
  author={Shaikh, Omar and Lam, Michelle and Hejna, Joey and Shao, Yijia and Cho, Hyundong and Bernstein, Michael and Yang, Diyi},
  booktitle={International Conference on Learning Representations},
  volume={2025},
  pages={20498--20525},
  year={2025}
}

@article{wang2026openclaw,
  title={Openclaw-rl: Train any agent simply by talking},
  author={Wang, Yinjie and Chen, Xuyang and Jin, Xiaolong and Wang, Mengdi and Yang, Ling},
  journal={arXiv preprint arXiv:2603.10165},
  year={2026}
}

@article{buening2026aligning,
  title={Aligning language models from user interactions},
  author={Buening, Thomas Kleine and H{\"u}botter, Jonas and P{\'a}sztor, Barna and Shenfeld, Idan and Ramponi, Giorgia and Krause, Andreas},
  journal={arXiv preprint arXiv:2603.12273},
  year={2026}
}

@inproceedings{sun2020test,
  title={Test-time training with self-supervision for generalization under distribution shifts},
  author={Sun, Yu and Wang, Xiaolong and Liu, Zhuang and Miller, John and Efros, Alexei and Hardt, Moritz},
  booktitle={International conference on machine learning},
  pages={9229--9248},
  year={2020},
  organization={PMLR}
}

@article{yuksekgonul2026learning,
  title={Learning to discover at test time},
  author={Yuksekgonul, Mert and Koceja, Daniel and Li, Xinhao and Bianchi, Federico and McCaleb, Jed and Wang, Xiaolong and Kautz, Jan and Choi, Yejin and Zou, James and Guestrin, Carlos and others},
  journal={arXiv preprint arXiv:2601.16175},
  year={2026}
}

@article{abdulhai2026llms,
  title={How llms distort our written language},
  author={Abdulhai, Marwa and White, Isadora and Wan, Yanming and Qureshi, Ibrahim and Leibo, Joel and Kleiman-Weiner, Max and Jaques, Natasha},
  journal={arXiv preprint arXiv:2603.18161},
  year={2026}
}

@article{rafailov2023direct,
  title={Direct preference optimization: Your language model is secretly a reward model},
  author={Rafailov, Rafael and Sharma, Archit and Mitchell, Eric and Manning, Christopher D and Ermon, Stefano and Finn, Chelsea},
  journal={Advances in neural information processing systems},
  volume={36},
  pages={53728--53741},
  year={2023}
}

@article{hu2022lora,
  title={Lora: Low-rank adaptation of large language models.},
  author={Hu, Edward J and Shen, Yelong and Wallis, Phillip and Allen-Zhu, Zeyuan and Li, Yuanzhi and Wang, Shean and Wang, Liang and Chen, Weizhu and others},
  journal={Iclr},
  volume={1},
  number={2},
  pages={3},
  year={2022}
}

@inproceedings{deutsch2022limitations,
  title={On the limitations of reference-free evaluations of generated text},
  author={Deutsch, Daniel and Dror, Rotem and Roth, Dan},
  booktitle={Proceedings of the 2022 Conference on Empirical Methods in Natural Language Processing},
  pages={10960--10977},
  year={2022}
}

@article{zheng2023judging,
  title={Judging llm-as-a-judge with mt-bench and chatbot arena},
  author={Zheng, Lianmin and Chiang, Wei-Lin and Sheng, Ying and Zhuang, Siyuan and Wu, Zhanghao and Zhuang, Yonghao and Lin, Zi and Li, Zhuohan and Li, Dacheng and Xing, Eric and others},
  journal={Advances in neural information processing systems},
  volume={36},
  pages={46595--46623},
  year={2023}
}

@inproceedings{shankar2024validates,
  title={Who validates the validators? aligning llm-assisted evaluation of llm outputs with human preferences},
  author={Shankar, Shreya and Zamfirescu-Pereira, JD and Hartmann, Bj{\"o}rn and Parameswaran, Aditya and Arawjo, Ian},
  booktitle={Proceedings of the 37th Annual ACM Symposium on User Interface Software and Technology},
  pages={1--14},
  year={2024}
}

@inproceedings{wortsman2022model,
  title={Model soups: averaging weights of multiple fine-tuned models improves accuracy without increasing inference time},
  author={Wortsman, Mitchell and Ilharco, Gabriel and Gadre, Samir Ya and Roelofs, Rebecca and Gontijo-Lopes, Raphael and Morcos, Ari S and Namkoong, Hongseok and Farhadi, Ali and Carmon, Yair and Kornblith, Simon and others},
  booktitle={International conference on machine learning},
  pages={23965--23998},
  year={2022},
  organization={Pmlr}
}

@article{cukurova2026you,
  title={What do you mean by human-AI collaboration: Prerequisite functions and the affordances needed to achieve it},
  author={Cukurova, Mutlu},
  journal={arXiv preprint arXiv:2606.15509},
  year={2026}
}

@article{xu2026mem,
  title={A-mem: Agentic memory for llm agents},
  author={Xu, Wujiang and Liang, Zujie and Mei, Kai and Gao, Hang and Tan, Juntao and Zhang, Yongfeng},
  journal={Advances in Neural Information Processing Systems},
  volume={38},
  pages={17577--17604},
  year={2026}
}

@article{wang2024voyager,
title={Voyager: An Open-Ended Embodied Agent with Large Language Models},
author={Guanzhi Wang and Yuqi Xie and Yunfan Jiang and Ajay Mandlekar and Chaowei Xiao and Yuke Zhu and Linxi Fan and Anima Anandkumar},
journal={Transactions on Machine Learning Research},
issn={2835-8856},
year={2024},
url={https://openreview.net/forum?id=ehfRiF0R3a},
note={}
}

@inproceedings{huang2024large,
  title={Large language models cannot self-correct reasoning yet},
  author={Huang, Jie and Chen, Xinyun and Mishra, Swaroop and Zheng, Huaixiu Steven and Yu, Adams and Song, Xinying and Zhou, Denny},
  booktitle={International conference on learning representations},
  volume={2024},
  pages={32808--32824},
  year={2024}
}

@inproceedings{xu2024pride,
  title={Pride and prejudice: LLM amplifies self-bias in self-refinement},
  author={Xu, Wenda and Zhu, Guanglei and Zhao, Xuandong and Pan, Liangming and Li, Lei and Wang, William},
  booktitle={Proceedings of the 62nd Annual Meeting of the Association for Computational Linguistics (Volume 1: Long Papers)},
  pages={15474--15492},
  year={2024}
}

@article{kojima2021continual,
    title = "Continual Learning for Grounded Instruction Generation by Observing Human Following Behavior",
    author = "Kojima, Noriyuki  and
      Suhr, Alane  and
      Artzi, Yoav",
    editor = "Roark, Brian  and
      Nenkova, Ani",
    journal = "Transactions of the Association for Computational Linguistics",
    volume = "9",
    year = "2021",
    address = "Cambridge, MA",
    publisher = "MIT Press",
    url = "https://aclanthology.org/2021.tacl-1.77/",
    doi = "10.1162/tacl_a_00428",
    pages = "1303--1319",
}

@inproceedings{hawkins2020continual,
    title = "Continual Adaptation for Efficient Machine Communication",
    author = "Hawkins, Robert  and
      Kwon, Minae  and
      Sadigh, Dorsa  and
      Goodman, Noah",
    editor = "Fern{\'a}ndez, Raquel  and
      Linzen, Tal",
    booktitle = "Proceedings of the 24th Conference on Computational Natural Language Learning",
    month = nov,
    year = "2020",
    address = "Online",
    publisher = "Association for Computational Linguistics",
    url = "https://aclanthology.org/2020.conll-1.33/",
    doi = "10.18653/v1/2020.conll-1.33",
    pages = "408--419",
}

@inproceedings{song2026expanding,
title={Expanding the Capabilities of Reinforcement Learning via Text Feedback},
author={Yuda Song and Lili Chen and Fahim Tajwar and R{\'e}mi Munos and Deepak Pathak and Drew Bagnell and Aarti Singh and Andrea Zanette},
booktitle={The 1st Workshop on Scaling Post-training for LLMs},
year={2026},
url={https://openreview.net/forum?id=om12baYnKM}
}

@article{weidinger2025toward,
  title={Toward an evaluation science for generative ai systems},
  author={Weidinger, Laura and Raji, Inioluwa Deborah and Wallach, Hanna and Mitchell, Margaret and Wang, Angelina and Salaudeen, Olawale and Bommasani, Rishi and Ganguli, Deep and Koyejo, Sanmi and Isaac, William},
  journal={arXiv preprint arXiv:2503.05336},
  year={2025}
}

@misc{cursor2026direct,
  title = {Direct Agents with Visual Prompts in Design Mode},
  author={Nilsson, Erik and Huang, Ian and Lu, Ryo},
  howpublished={\url{https://cursor.com/blog/design-mode}},
  year="2026",
}

@inproceedings{ross2011reduction,
  title={A reduction of imitation learning and structured prediction to no-regret online learning},
  author={Ross, St{\'e}phane and Gordon, Geoffrey and Bagnell, Drew},
  booktitle={Proceedings of the fourteenth international conference on artificial intelligence and statistics},
  pages={627--635},
  year={2011},
  organization={JMLR Workshop and Conference Proceedings}
}

@inproceedings{park2024rlhf,
title={{RLHF} from Heterogeneous Feedback via Personalization and Preference Aggregation},
author={Chanwoo Park and Mingyang Liu and Dingwen Kong and Kaiqing Zhang and Asuman E. Ozdaglar},
booktitle={ICML 2024 Workshop on Theoretical Foundations of Foundation Models},
year={2024},
url={https://openreview.net/forum?id=8XI7AGByAp}
}

@inproceedings{chakraborty2024maxminrlhf,
title={MaxMin-{RLHF}: Alignment with Diverse Human Preferences},
author={Souradip Chakraborty and Jiahao Qiu and Hui Yuan and Alec Koppel and Dinesh Manocha and Furong Huang and Amrit Bedi and Mengdi Wang},
booktitle={Forty-first International Conference on Machine Learning},
year={2024},
url={https://openreview.net/forum?id=8tzjEMF0Vq}
}

@inproceedings{liu2023geval,
    title = "{G}-Eval: {NLG} Evaluation using Gpt-4 with Better Human Alignment",
    author = "Liu, Yang  and Iter, Dan  and Xu, Yichong  and Wang, Shuohang  and Xu, Ruochen  and Zhu, Chenguang",
    booktitle = "Proceedings of the 2023 Conference on Empirical Methods in Natural Language Processing",
    month = dec,
    year = "2023",
    address = "Singapore",
    publisher = "Association for Computational Linguistics",
    url = "https://aclanthology.org/2023.emnlp-main.153/",
    doi = "10.18653/v1/2023.emnlp-main.153",
    pages = "2511--2522",
}

@article{tang2024understanding,
  title={Understanding the performance gap between online and offline alignment algorithms},
  author={Tang, Yunhao and Guo, Daniel Zhaohan and Zheng, Zeyu and Calandriello, Daniele and Cao, Yuan and Tarassov, Eugene and Munos, R{\'e}mi and Pires, Bernardo {\'A}vila and Valko, Michal and Cheng, Yong and others},
  journal={arXiv preprint arXiv:2405.08448},
  year={2024}
}

@inproceedings{
zhao2026selfdistilled,
title={Self-Distilled Reasoner: On-Policy Self-Distillation for Large Language Models},
author={Siyan Zhao and Zhihui Xie and Mengchen Liu and Jing Huang and Guan Pang and Feiyu Chen and Aditya Grover},
booktitle={ICLR 2026 Workshop on Lifelong Agents: Learning, Aligning, Evolving},
year={2026},
url={https://openreview.net/forum?id=31wkVZXEaJ}
}

@article{tiwari2026learning,
  title={Learning, fast and slow: Towards LLMs that adapt continually},
  author={Tiwari, Rishabh and Sareen, Kusha and Agrawal, Lakshya A and Gonzalez, Joseph E and Zaharia, Matei and Keutzer, Kurt and Dhillon, Inderjit S and Agarwal, Rishabh and Khatri, Devvrit},
  journal={arXiv preprint arXiv:2605.12484},
  year={2026}
}
